\documentclass[aps,pra,superscriptaddress,twocolumn,longbibliography]{revtex4-1}
\usepackage{mathrsfs}
\usepackage{amsfonts}
\usepackage{amssymb}
\usepackage{amsmath}
\usepackage{graphicx}
\usepackage{lipsum} 
\usepackage{ulem}
\usepackage{sidecap}
\usepackage{color}
\usepackage{tikz}
\usepackage[colorlinks=true, urlcolor=blue, citecolor=blue,linkcolor=blue,citebordercolor={1 0 0},linkbordercolor={0 0 1}]{hyperref}
\usepackage{subeqnarray}
\usepackage{verbatim}
\usepackage{euscript} 

\usepackage{shortcuts}

\usepackage{array}
\usepackage{multirow}

\renewcommand{\eqref}[1]{Eq.~(\ref{#1})} 

\definecolor{mydarkblue}{rgb}{0,0.08,0.45}
\definecolor{myfavblue}{rgb}{0.1176, 0.392, 1.0}
\hypersetup{
    colorlinks=true,
    linkcolor=mydarkblue,
    citecolor=mydarkblue,
    filecolor=mydarkblue,
    urlcolor=mydarkblue}

\newcommand{\utchem}{Department  of  Chemistry,  University  of  Toronto,  Toronto,  Ontario  M5G 1Z8,  Canada}
\newcommand{\utcomp}{Department  of  Computer Science,  University  of  Toronto,  Toronto,  Ontario  M5S 2E4,  Canada}
\newcommand{\vectorinst}{Vector  Institute  for  Artificial  Intelligence,  Toronto,  Ontario  M5S  1M1,  Canada}
\newcommand{\cifar}{Canadian  Institute  for  Advanced  Research,  Toronto,  Ontario  M5G  1Z8,  Canada}

\begin{document}

\title{Keeping it Simple: Language Models can learn Complex Molecular Distributions}

\author{Daniel Flam-Shepherd}
\email{danielfs@cs.toronto.edu}
\affiliation{\utcomp}
\affiliation{\vectorinst}

\author{Kevin Zhu}
\affiliation{\utcomp}

\author{Al\'an Aspuru-Guzik}
\email{alan@aspuru.com}
\affiliation{\utcomp}
\affiliation{\vectorinst}
\affiliation{\utchem}
\affiliation{\cifar}

\date{\today}

\begin{abstract}
Deep generative models of molecules have grown immensely in popularity, trained on relevant datasets, these models are used to search through chemical space. 
The downstream utility of generative models for the inverse design of novel functional compounds, depends on their ability to learn a training distribution of molecules. 
The most simple example is a language model that takes the form of a recurrent neural network and generates molecules using a string representation.
More sophisticated are graph generative models, which sequentially construct molecular graphs and typically achieve state of the art results. 
However, recent work has shown that language models are more capable than once thought, particularly in the low data regime.  
In this work, we investigate the capacity of simple language models to learn distributions of molecules. 
For this purpose, we introduce several challenging generative modeling tasks by compiling especially complex distributions of molecules.
On each task, we evaluate the ability of language models as compared with two widely used graph generative models.
The results demonstrate that language models are powerful generative models, capable of adeptly learning complex molecular distributions--  and yield better performance than the graph models. 
Language models can accurately generate: distributions of the highest scoring penalized LogP molecules in ZINC15, multi-modal molecular distributions as well as the largest molecules in PubChem. 
\end{abstract}

\maketitle


The efficient exploration of chemical space is one of the most important objectives in all of science, 
with numerous applications in therapeutics and materials discovery. 
However, exploration efforts have only probed a very small subset of the synthetically accessible chemical space \cite{bohacek1996art}, 
therefore developing new tools is essential. It is possible that the rise of artificial intelligence will provide the methods to unlock the mysteries of the chemical universe, given its success in other challenging scientific questions like protein structure prediction \cite{jumper2021highly}. 

Very recently, deep generative models have emerged as one of the most promising tool for this immense challenge \cite{gomez2018automatic}.
These models are trained on relevant subsets of chemical space and can generate novel molecules similar to their training data. 
Their ability to learn the training distribution and generate valid, similar molecules-- is important for success in downstream applications like the inverse design of functional compounds. 

The first models involved re-purposing recurrent neural networks (RNNs) \cite{sutskever2011generating} in order to generate molecules as SMILES strings \cite{weininger1988smiles}. 
These language models can be used to generate molecular libraries for drug discovery \cite{segler2018generating} or built into variational autoencoders (VAE) \cite{kingma2013auto, gomez2018automatic} where bayesian optimization can be used to search through the model's latent space for drug-like molecules. 
However there are problems with language models that make it challenging to train models' capable of generating valid molecules. 
The largest issue is the brittleness of the SMILES string representation that is widely used in these models. 
If a single character is misplaced or erroneously generated then the produced SMILES string will correspond to an invalid molecule. 
This makes it difficult to train and apply language models for practical applications, in spite of this, researchers have attempted to use them for ligand based de novo design \cite{perron2021deep}. 

Other models that use graph representations do not have this issue and can be directly constrained to enforce valency restrictions. 
These graph generative models sequentially construct molecules as a series of probabilistic decisions 
\cite{li2018learning,liu2018constrained,jin2018junction, you2018graph,seff2019discrete,samanta2019nevae} 
using graph neural networks \cite{DuvMacetal15nfp, flam2021neural}. 
These models have been shown to be more capable in modeling distributions and useful in the inverse design of molecules with specific properties \cite{jin2018junction, you2018graph}. 
Most importantly, they will always generate valid molecules in contrast to SMILES language models and others that generate whole molecules in one shot 
\cite{simonovsky2018graphvae,ma2018constrained, de2018molgan, flam2020graph}. 


Recently, more robust molecular string representations have been proposed \cite{kusner2017grammar,dai2018syntax, o2018deepsmiles}. In particular, self referencing embedding strings (SELFIES) \cite{krenn2019selfies} distinguish themselves as every SELFIES string is a valid molecule. Additionally, with improved training methods, language models have achieved promising results \cite{liu2018constrained,polykovskiy2020molecular}, particularly in the low-data regime \cite{skinnider2021deep,moret2020generative}. 
Given these developments, in order to test the ability of language models, we formulate a series of difficult generative modeling tasks by constructing training sets of molecules that are especially challenging to learn. 
We train language models on all tasks and provide baseline comparisons with graph models.
The results demonstrate that language models are powerful generative models and can learn very complex molecular distributions that some popular graph models are less able or completely unable to.

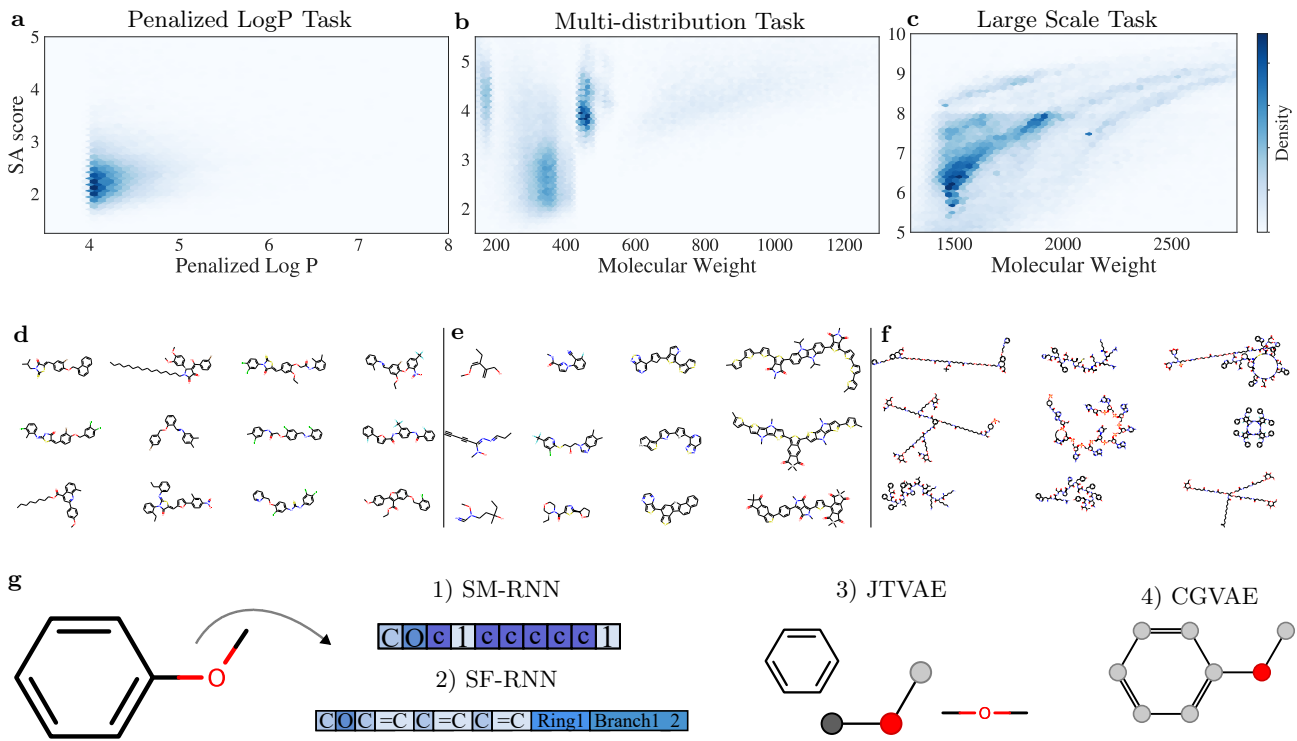
\begin{figure*}[t]
\begin{tikzpicture}
    \draw (0, 0) node[inner sep=0] {\includegraphics[width=0.33\textwidth]{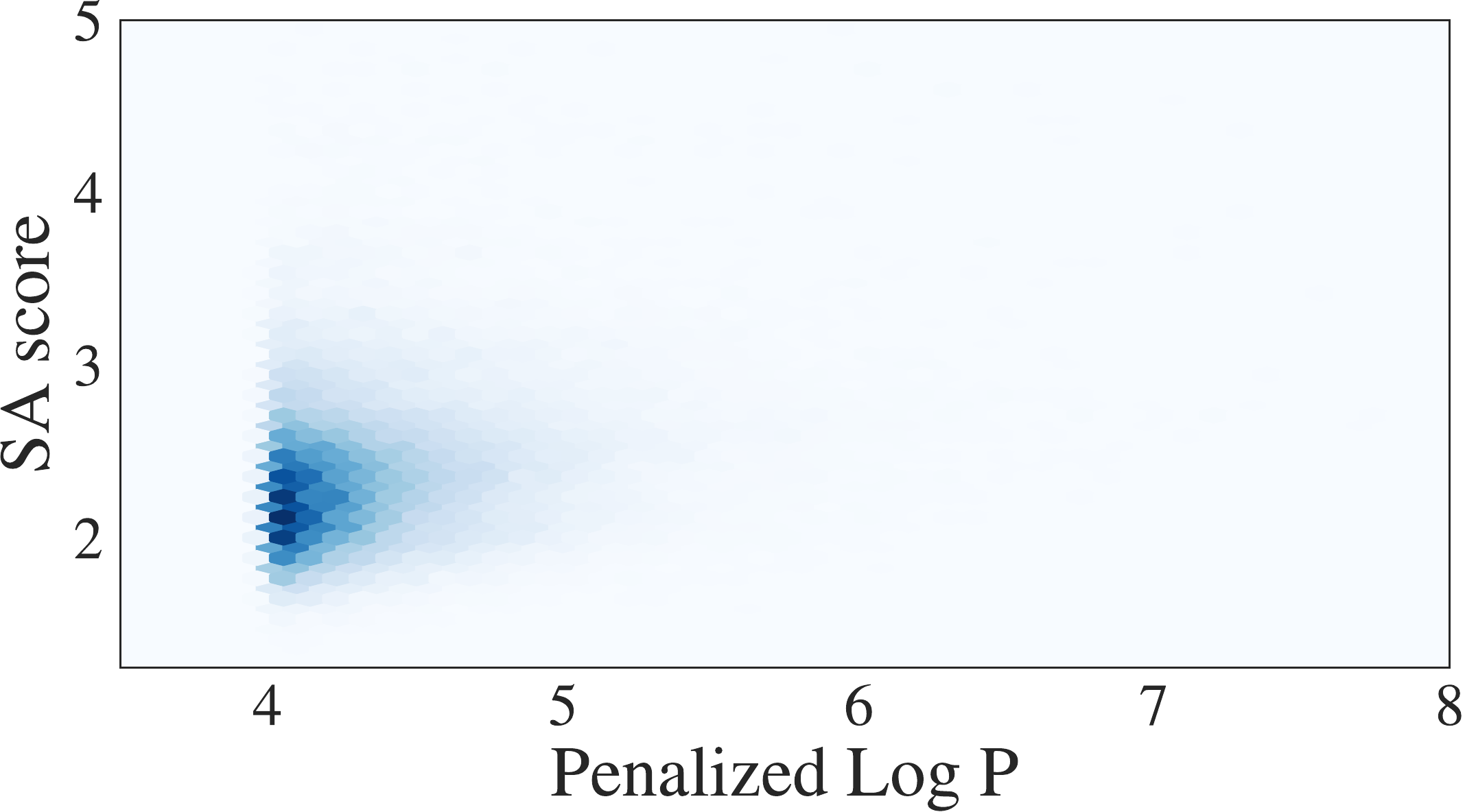} 
                                    \includegraphics[width=0.31\textwidth]{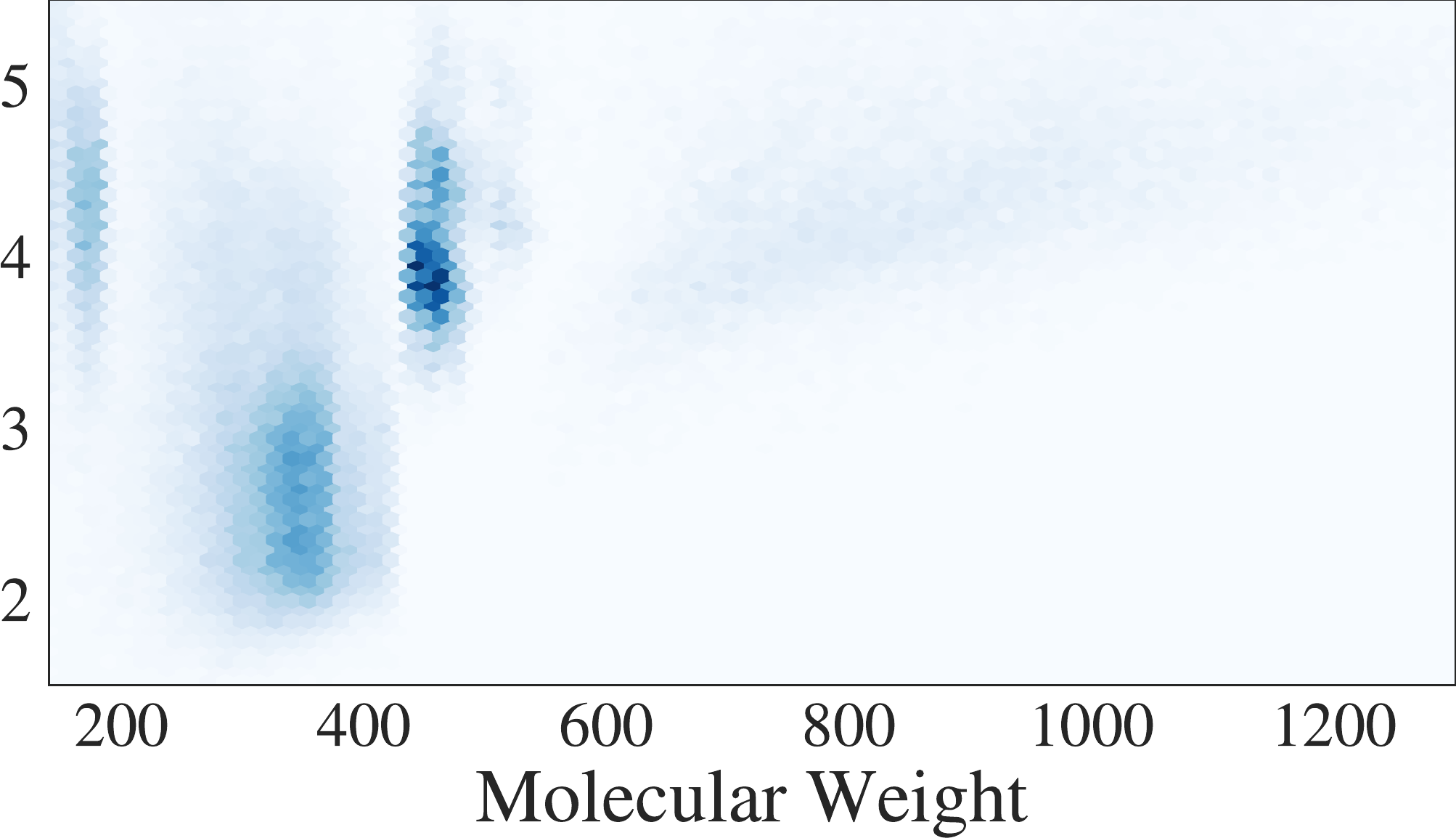} 
                                    \includegraphics[width=0.3\textwidth]{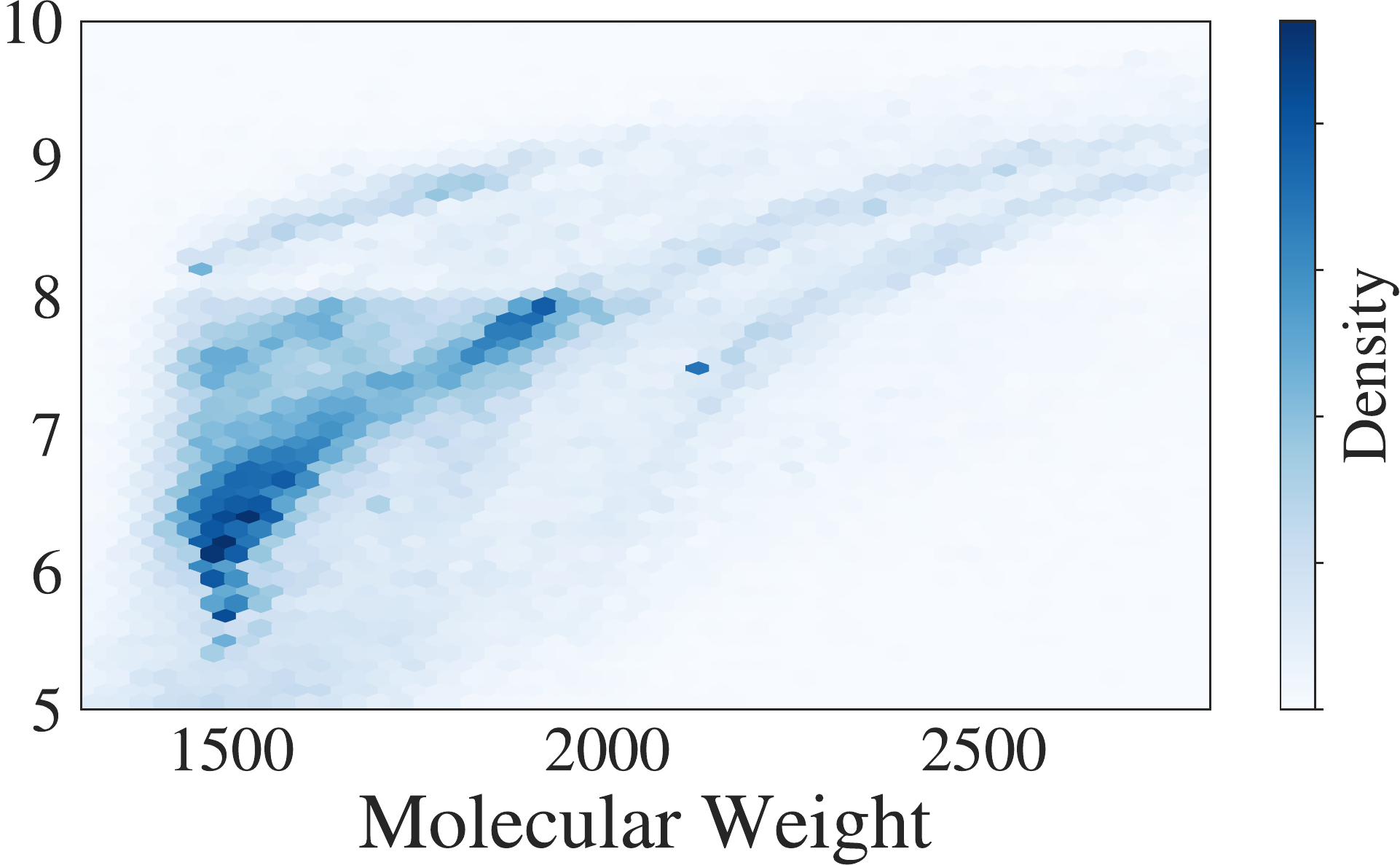}};
    \draw (-8.4, 1.75) node[] {\textbf{a}};
    \draw (-2.5, 1.75) node[] {\textbf{b}};
    \draw (3.5, 1.75) node[] {\textbf{c}};
    \draw (-5.45, 1.75) node[] {Penalized LogP Task};
    \draw (0.4, 1.75) node[] {Multi-distribution Task};
    \draw (5.55, 1.75) node[] {Large Scale Task};
\end{tikzpicture}

\vspace{0.5cm}

\begin{tikzpicture}
    \draw (0, 0) node[inner sep=0] { \includegraphics[width=0.31\textwidth]{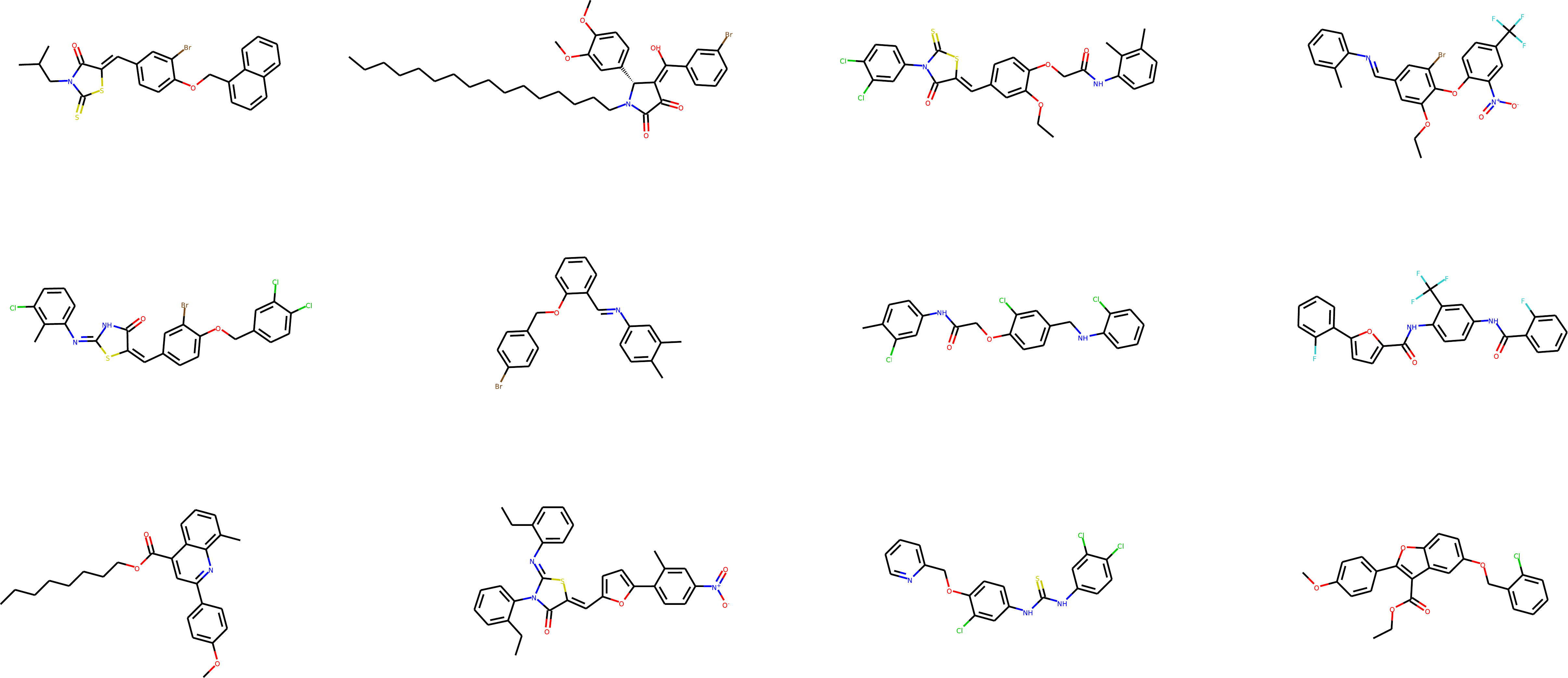} \vline 
                                     \includegraphics[width=0.31\textwidth]{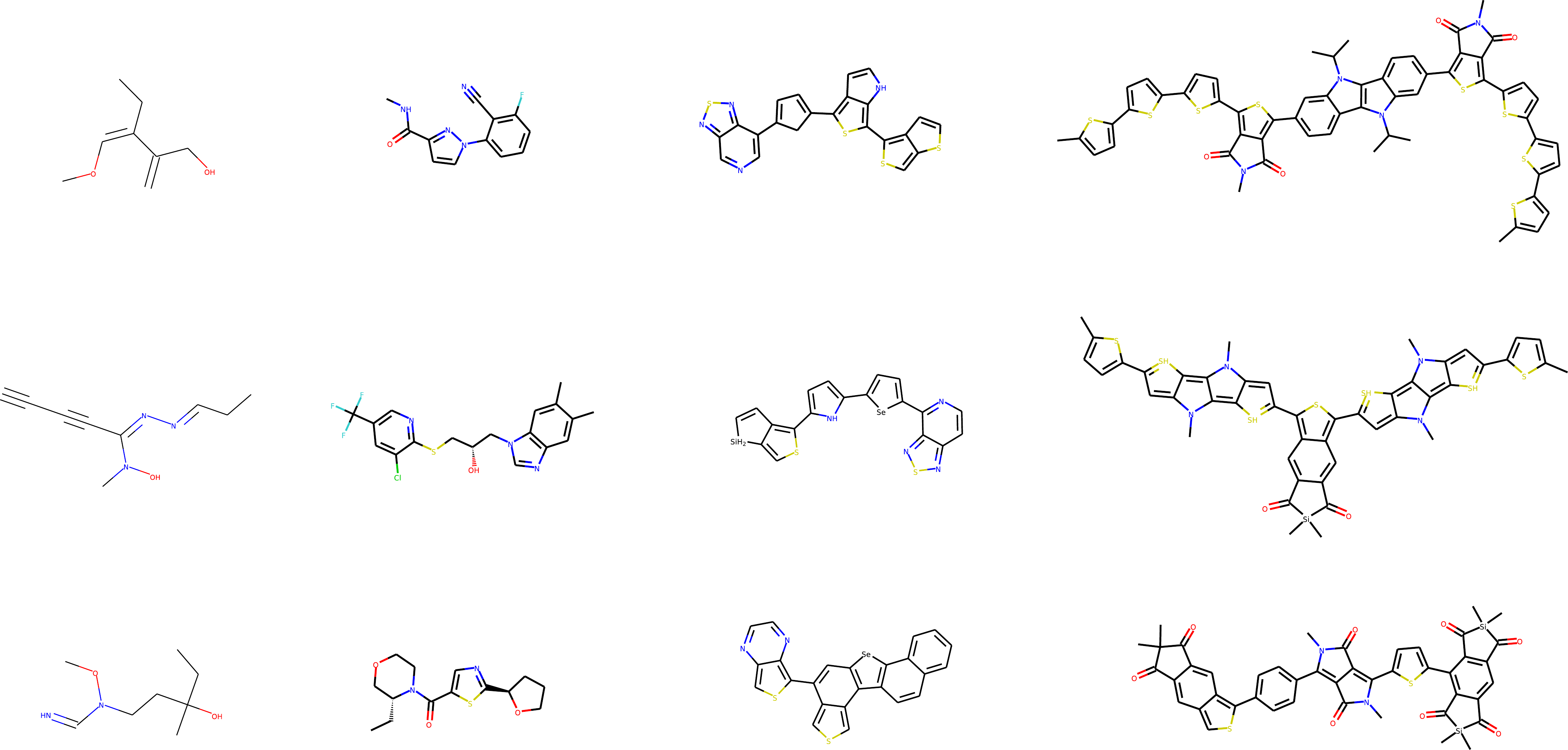} \vline 
                                     \includegraphics[width=0.31\textwidth]{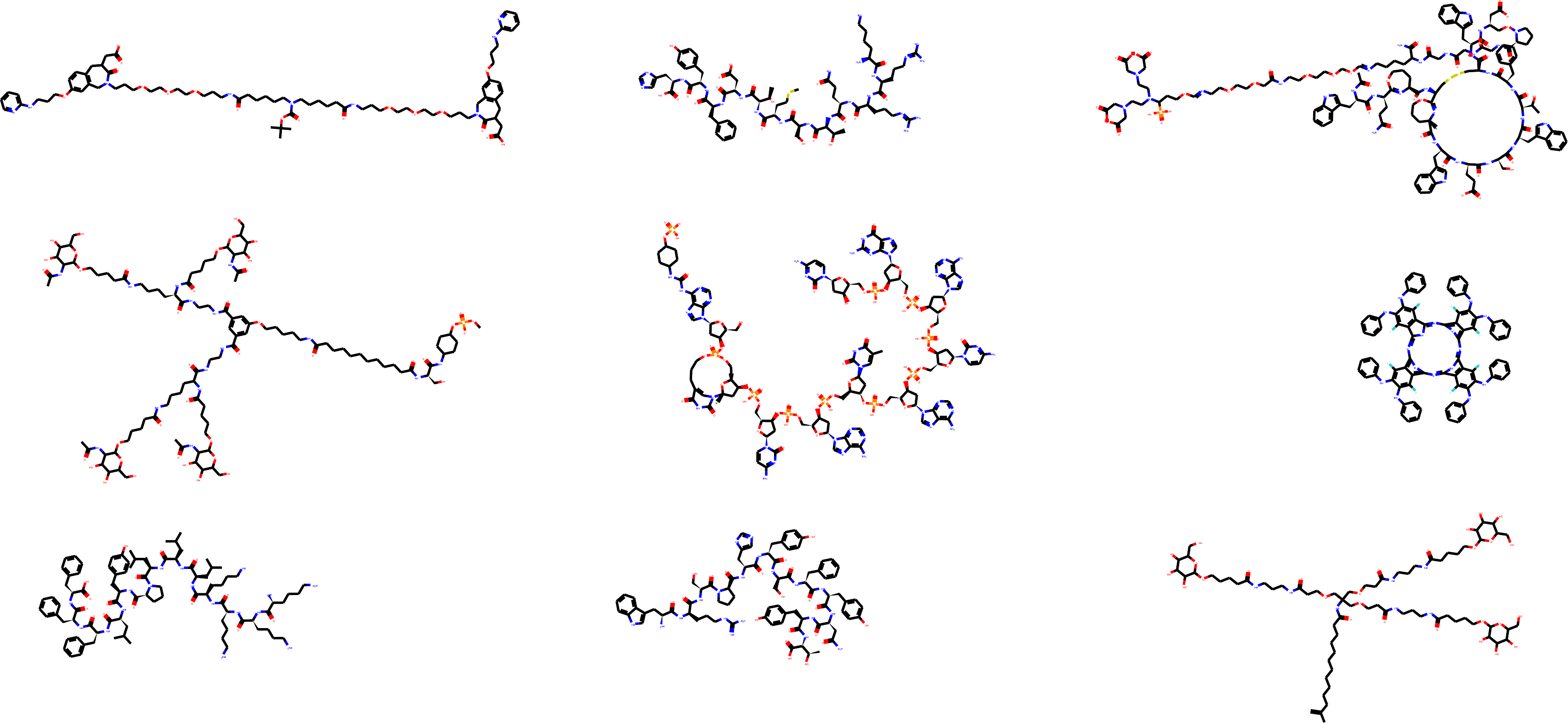}   };
    \draw (-8.4, 1.25) node[] {\textbf{d}};
    \draw (-2.6, 1.25) node[] {\textbf{e}};
    \draw (3.1, 1.25) node[] {\textbf{f}};
\end{tikzpicture}

\vspace{0.5cm}

\begin{tikzpicture}
    \draw (0, 0) node[inner sep=0] { \includegraphics[width=0.95\textwidth]{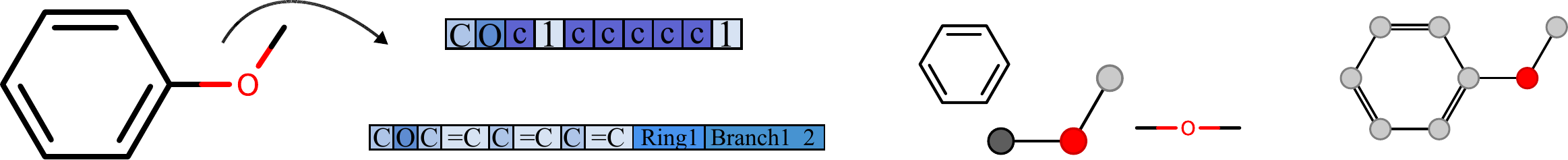} };
    \draw (-8.5, 1.2) node[] {\textbf{g}};
    \draw (-2.1, 1.1) node[] {1) SM-RNN};
   \draw (-2.1, -0.10) node[] {2) SF-RNN};
    \draw (3.15, 1.1) node[] {3) JTVAE};
    \draw (7.25, 1.) node[] {4) CGVAE};
\end{tikzpicture}
\caption{\textbf{a-c} The molecular distributions defining the three complex molecular generative modeling task.
         \textbf{a} The distribution of penalized LogP vs SA score from the training data in the penalized logP task.
         \textbf{b} The four modes of differently weighted molecules in the training data of the multi-distribution task
         \textbf{c} Large scale task's molecular weight training distribution.
         \textbf{d-f} examples of molecules plotted using rdkit \cite{landrum2013rdkit} from the training data in each of the generative modeling tasks 
         \textbf{d} The penalized LogP task \textbf{e} The multi-distribution task \textbf{f} The large scale task
         \textbf{g} The main models considered and the representation they learn on: 1) SM-RNN trained on SMILES strings 2) SF-RNN trained on SELFIES 3) JTVAE using a graph and tree representation and 4) CGAVE using a graph representation.}
\label{fig:tasks}
\end{figure*}

\begin{figure*}[t]

\begin{tikzpicture}
    \draw (0, 0) node[inner sep=0] {\includegraphics[width=0.99\textwidth]{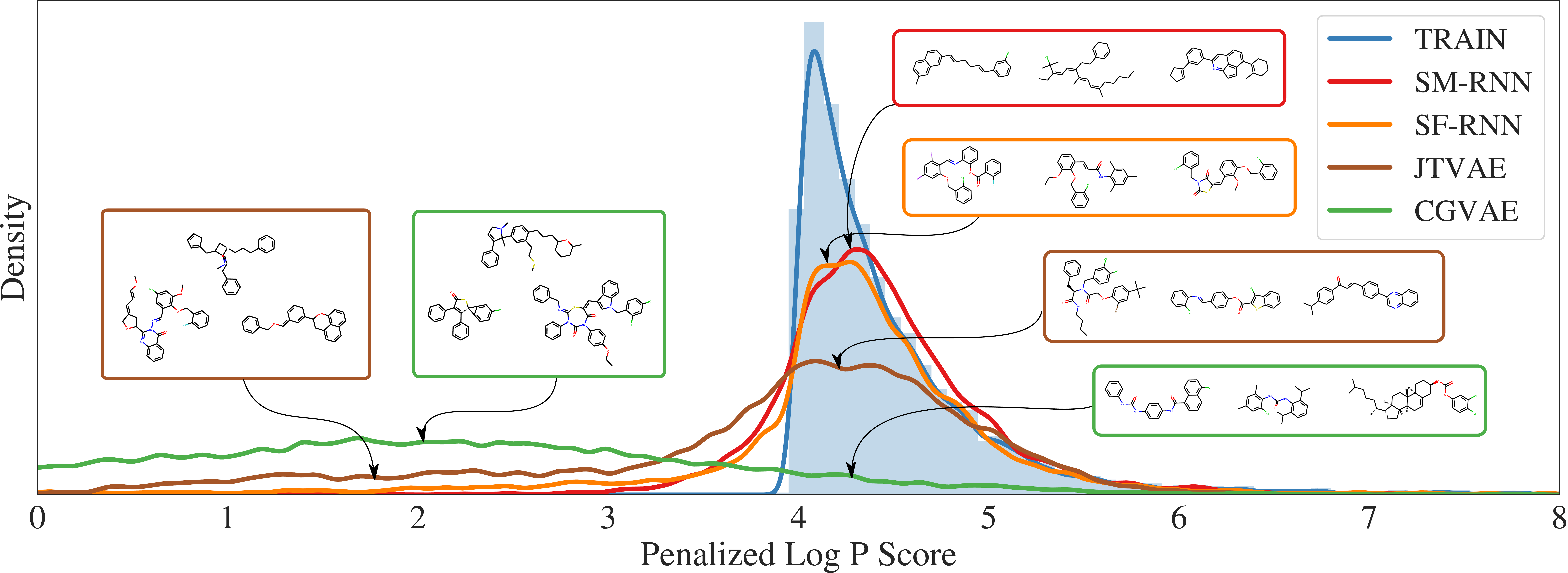} };
    \draw (-8.7, 3.1) node[] {\textbf{a}};
\end{tikzpicture}

\vspace{0.5cm}

\begin{tikzpicture}
    \draw (0, 0) node[inner sep=0] {\includegraphics[width=0.99\textwidth]{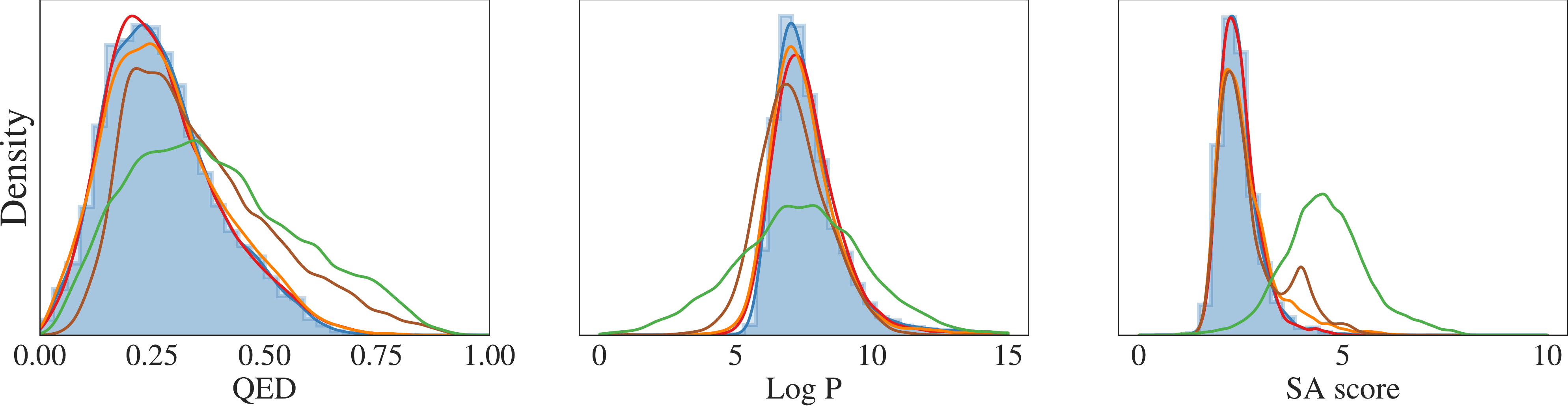} };
    \draw (-8.7, 2.1) node[] {\textbf{b}};
\end{tikzpicture}

\caption{\textbf{Penalized LogP Task} \ \textbf{a} The plotted distribution of the penalized LogP scores of molecules from the training data (TRAIN) with the SM-RNN trained on SMILES, the SF-RNN trained on SELFIES and graph models: CGVAE and JTVAE. 
For the graph models we display 3 molecules from the out of distribution mode at penalized LogP score $\in [1.5,2.5]$ as well as 3 molecules with penalized LogP score in the the main mode [4.0,4.5] from all models. 
\ \ \textbf{b} Distribution plots for all models and training data of molecular properties QED, LogP and SA score.}
\label{fig:logp}
\end{figure*}

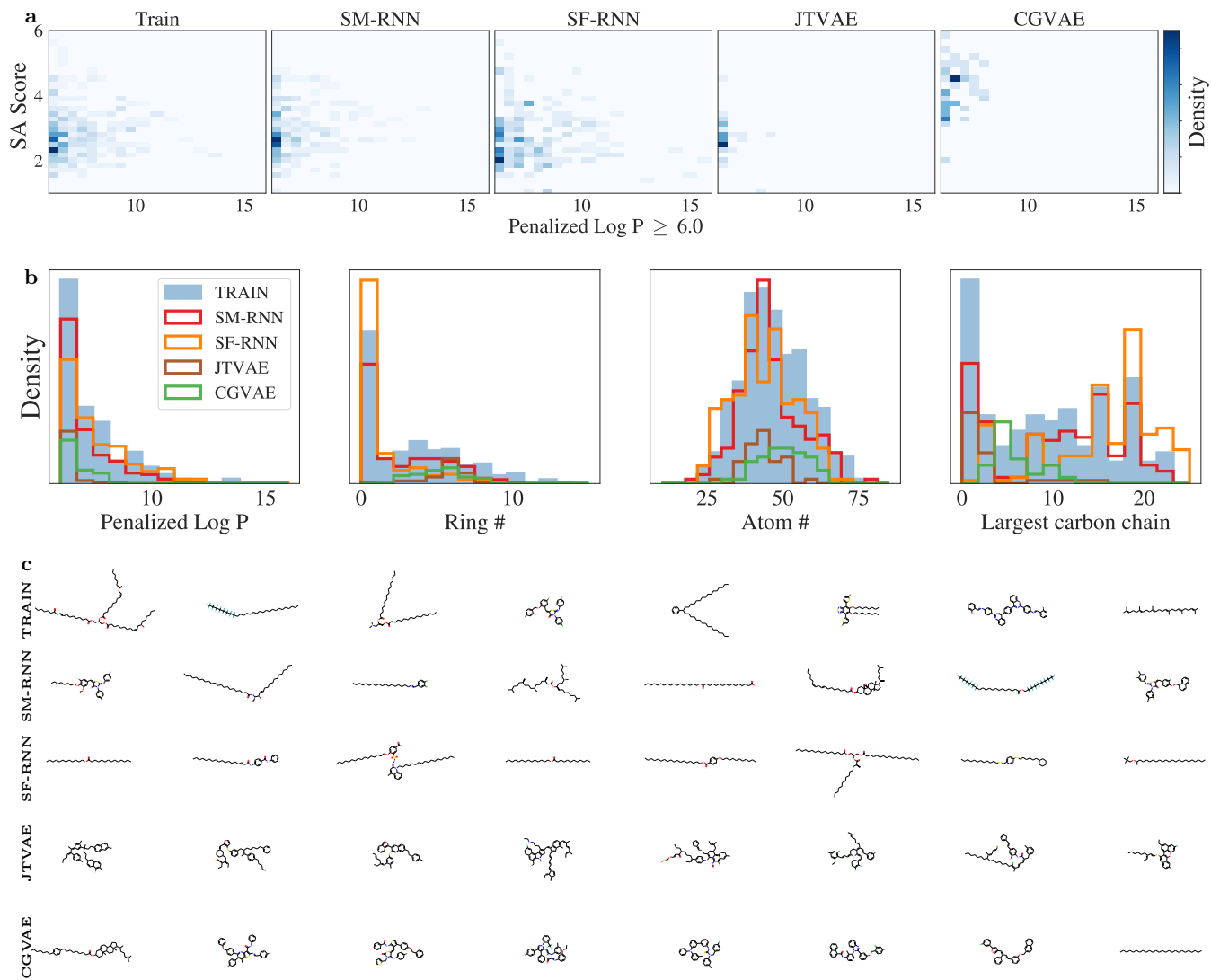
\begin{figure*}[t]

\begin{tikzpicture}
    \draw (0, 0) node[inner sep=0] {\includegraphics[width=0.99\textwidth]{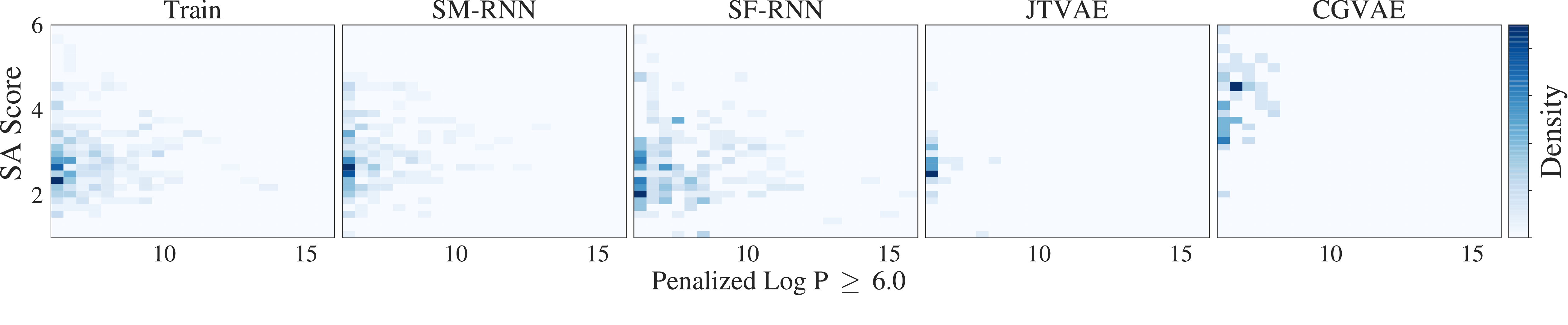}  };
    \draw (-8.56, 1.75) node[] {\textbf{a}};
\end{tikzpicture}

 \begin{tikzpicture}
    \draw (0, 0) node[inner sep=0] {\includegraphics[width=0.999\textwidth]{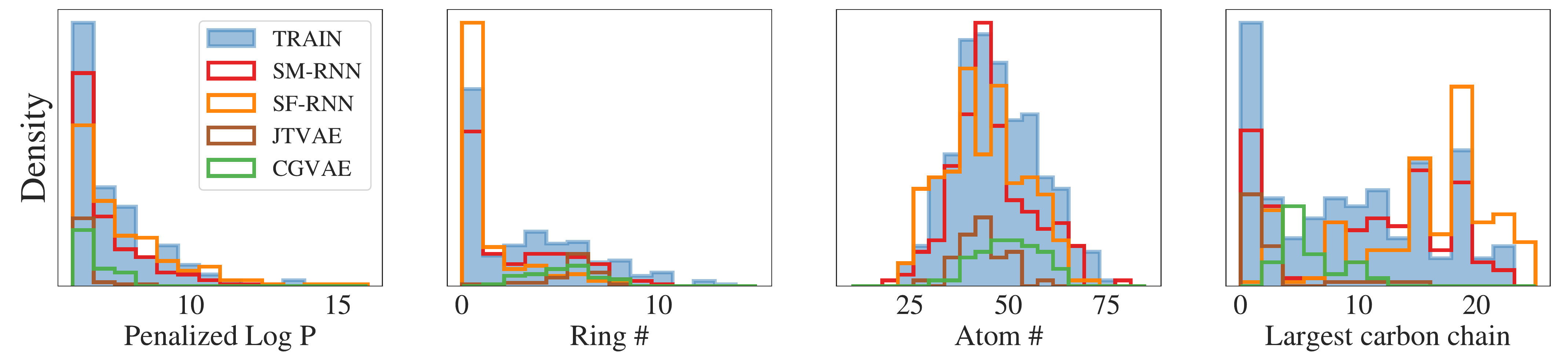}  };
    \draw (-8.56, 1.8) node[] {\textbf{b}};
\end{tikzpicture}

\vspace{0.25cm}

 \begin{tikzpicture}
    \draw (0, 0) node[inner sep=0] {\includegraphics[width=0.97\textwidth]{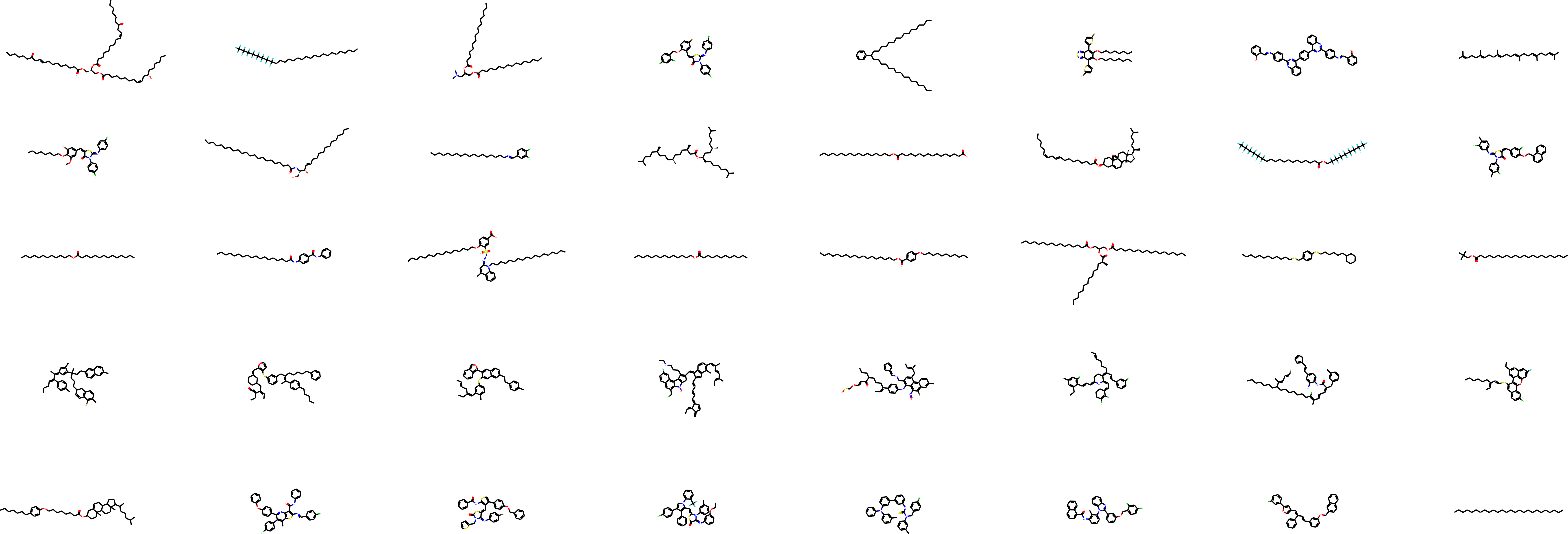}  };
    \draw (-8.75, 3.01) node[] {\textbf{c}};
    \draw (-8.75, 2.35) node[rotate=90] {\tiny\textbf{TRAIN}};
    \draw (-8.75, 1.225)  node[rotate=90] {\tiny\textbf{SM-RNN}};
    \draw (-8.75, 0.0) node[rotate=90] {\tiny\textbf{SF-RNN}};
    \draw (-8.75, -1.25)  node[rotate=90] {\tiny\textbf{JTVAE}};
    \draw (-8.75, -2.65) node[rotate=90] {\tiny\textbf{CGVAE}};
\end{tikzpicture}

\caption{\textbf{Penalized LogP Task} \ \textbf{a} 2d histograms of penalized LogP and SA score from molecules generated by the models or from training data that have penalized LogP $\geq 6.0$. \ \textbf{b} Histograms of penalized LogP, Atoms \#, Ring \# and length of largest carbon chain (all per molecule) from molecules generated by all models or from the training data that have penalized LogP $\geq 6.0$. \ \textbf{c} 10 molecules generated by the models or from the training data that have penalized LogP $\geq 6.0$. }
\label{fig:hlogp}
\end{figure*}

\begin{figure*}[t]

 \begin{tikzpicture}
    \draw (0, 0) node[inner sep=0] {\includegraphics[width=0.99\textwidth]{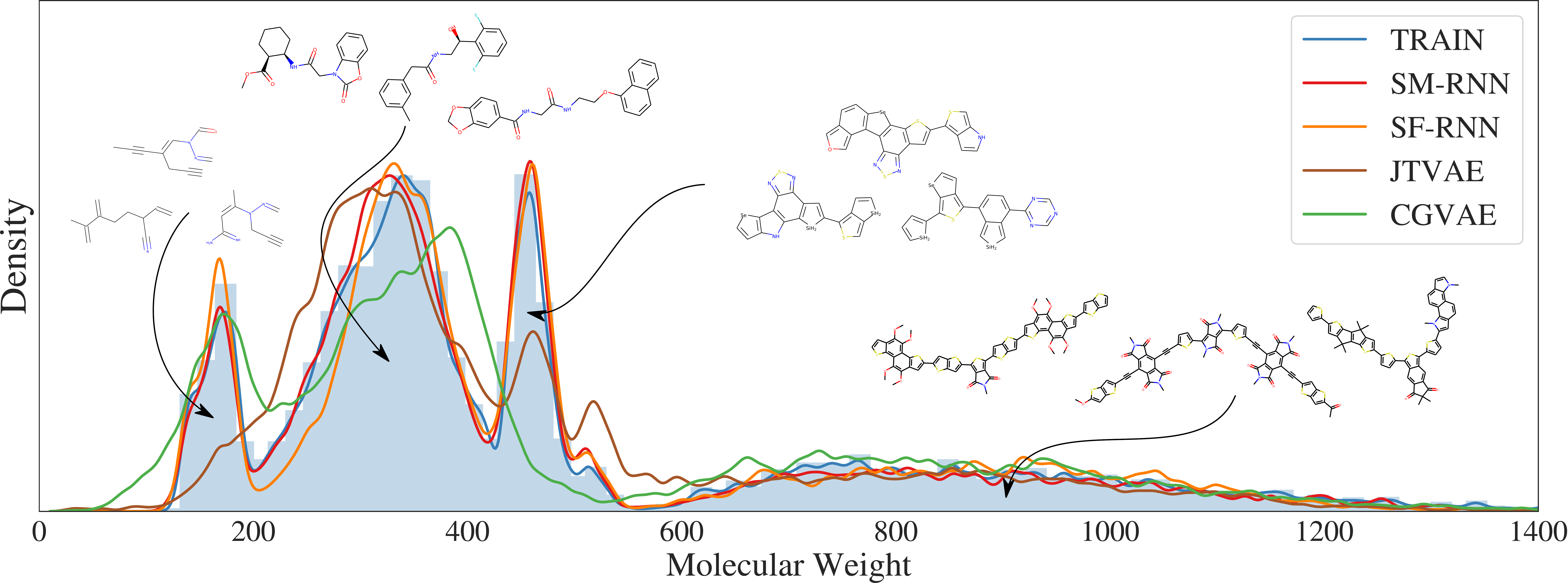}};
    \draw (-8.67, 3.15) node[] {\textbf{a}};
\end{tikzpicture}

\vspace{0.125cm}

\raggedright

\begin{tikzpicture}
    \draw (0, 0) node[inner sep=0] {\includegraphics[width=0.97\textwidth]{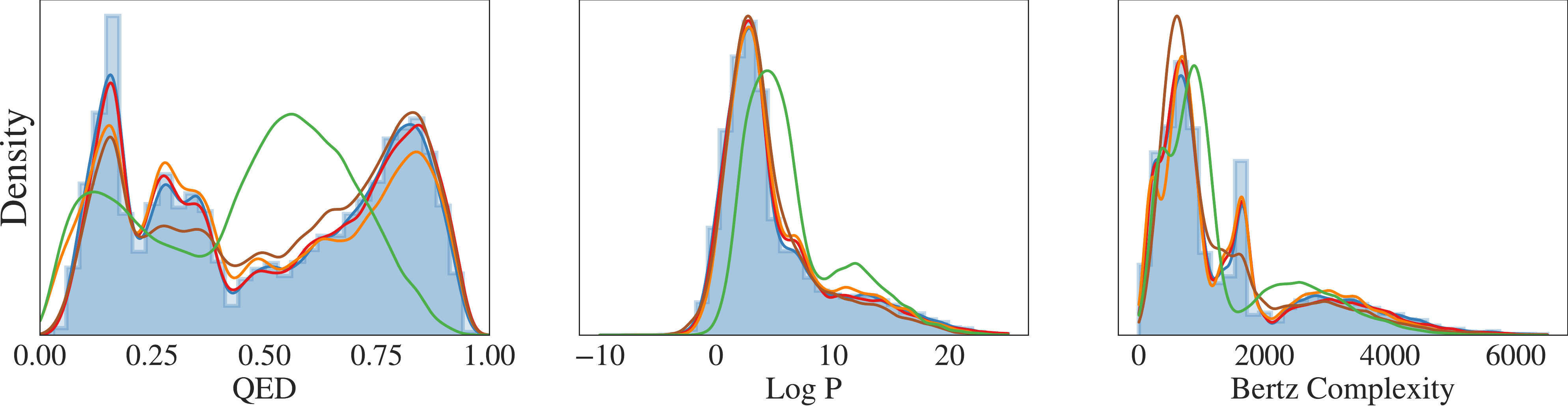}};
    \draw (-8.5, 2.12) node[] {\textbf{b}};
\end{tikzpicture}

\centering

\vspace{0.25cm}

\begin{tikzpicture}
    \draw (0, 0) node[inner sep=0] {\includegraphics[width=0.99\textwidth]{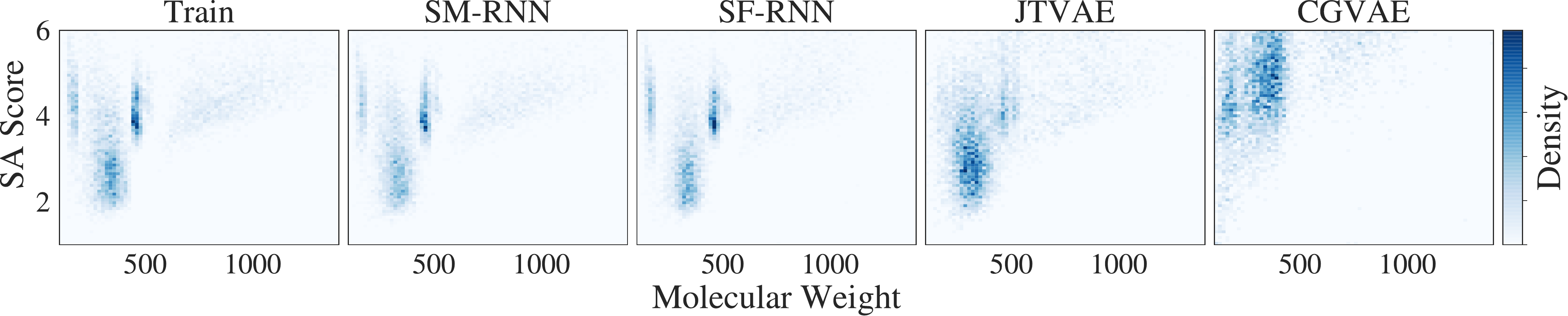}};
    \draw (-8.7, 1.558) node[] {\textbf{c}};
\end{tikzpicture}

\vspace{-0.2cm}

\caption{\textbf{Multi-distribution Task} 
\textbf{a} The Histogram and KDE of molecular weight of training molecules along with KDEs of molecular weight of molecules generated from all models.
Three training molecules from each mode are shown. 
\ \textbf{b} The Histogram and KDE of QED, LogP and SA scores of training molecules along with KDES of molecules generated from all models. 
\ \textbf{c} 2d histograms of molecular weight and SA score of training molecules and molecules generated by all models.}
\label{fig:md}
\end{figure*}
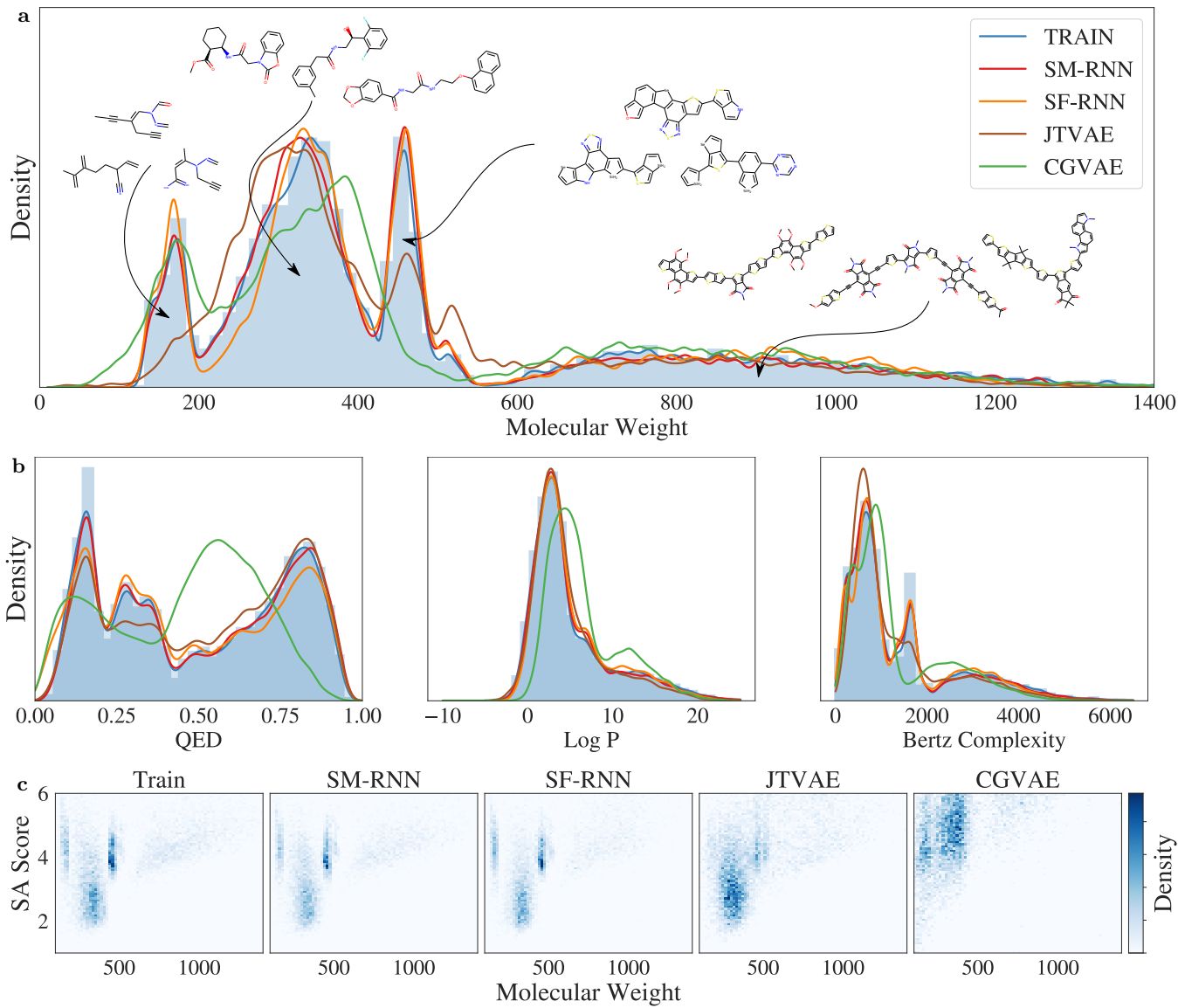

\begin{figure*}[t]

\begin{tikzpicture}
    \draw (0, 0) node[inner sep=0] {\includegraphics[width=0.999\textwidth]{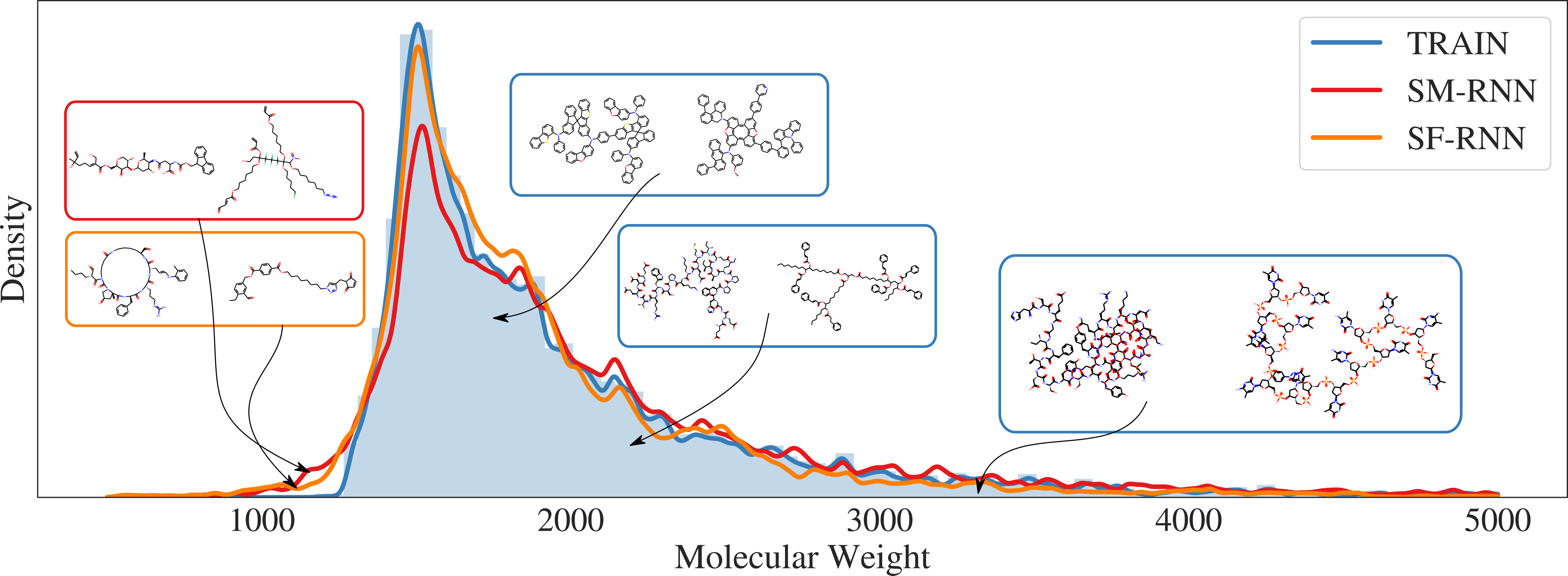}};
    \draw (-8.7, 3.1) node[] {\textbf{a}};
\end{tikzpicture}

\vspace{0.25cm}

\begin{tikzpicture}
    \draw (0, 0) node[inner sep=0] {\includegraphics[width=0.99\textwidth]{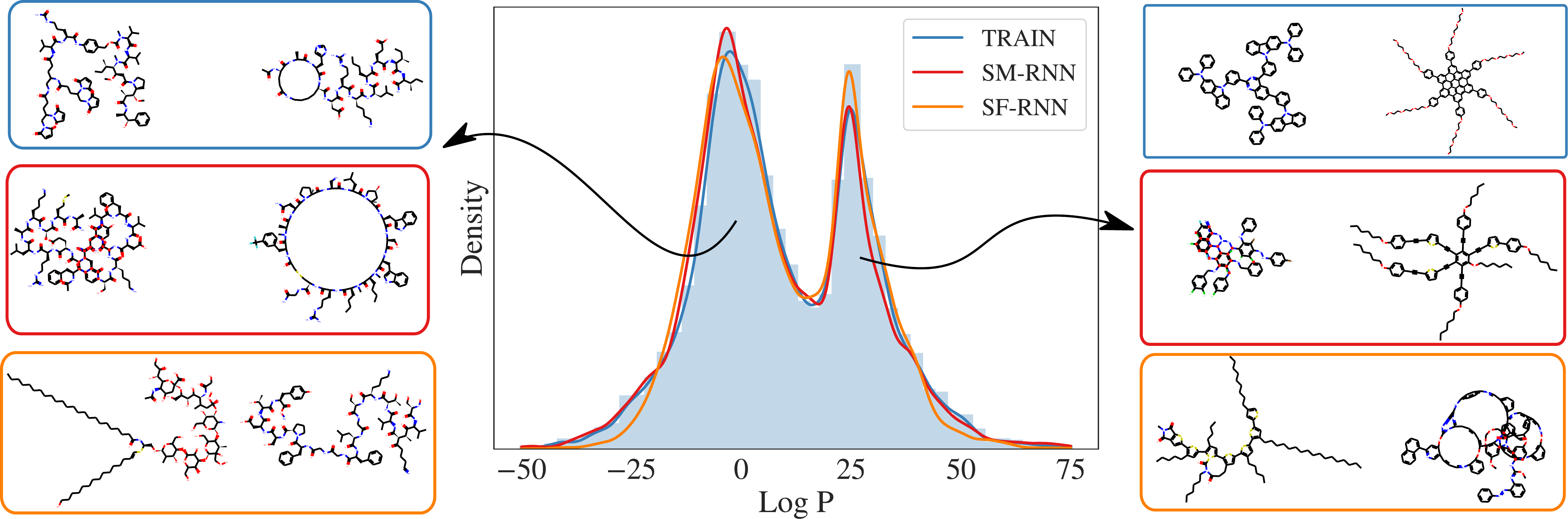}};
    \draw (-8.7, 3.15) node[] {\textbf{b}};
\end{tikzpicture}

\caption{\textbf{Large Scale Task} \ \textbf{a} The Histogram and KDE of molecular weight of training molecules along with the KDEs of molecular weight of molecules generated from the RNNs. 
Two molecules generated by the RNN's with lower molecular weight than most training molecules are shown on the left of the plot. In addition, two training molecules from three different regions in the distribution of molecular weight are displayed. \ \textbf{b} The Histogram and KDE of LogP of training molecules along with the KDEs of LogP of molecules generated from the RNNs. On either side of the plot, for each mode in the LogP distribution, we display 2 molecules from the training data and generated by the RNNs.}
\label{fig:ls}
\end{figure*}

\begin{figure*}[t]

\begin{tikzpicture}
    \draw (0, 0) node[inner sep=0] {\includegraphics[width=0.99\textwidth]{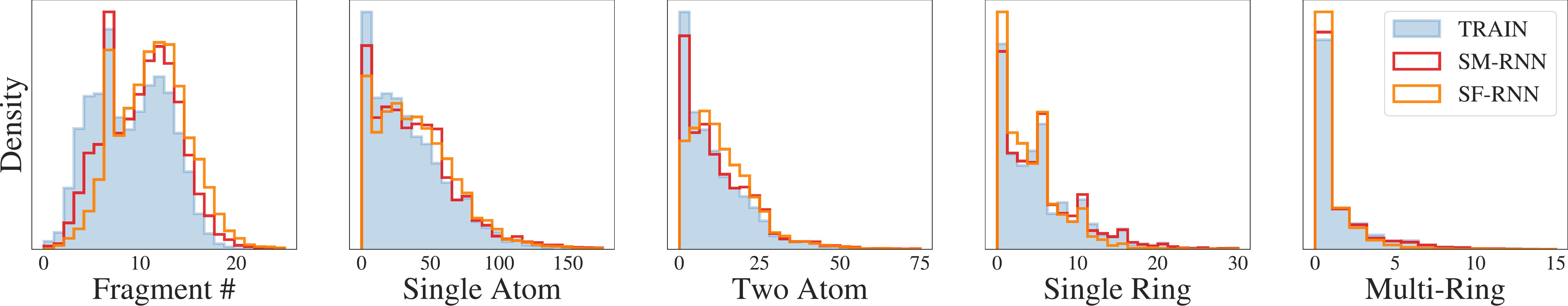}};
    \draw (-8.75, 1.55) node[] {\textbf{a}};
\end{tikzpicture}

\vspace{0.25cm}

\begin{tikzpicture}
    \draw (0, 0) node[inner sep=0] {\includegraphics[width=0.99\textwidth]{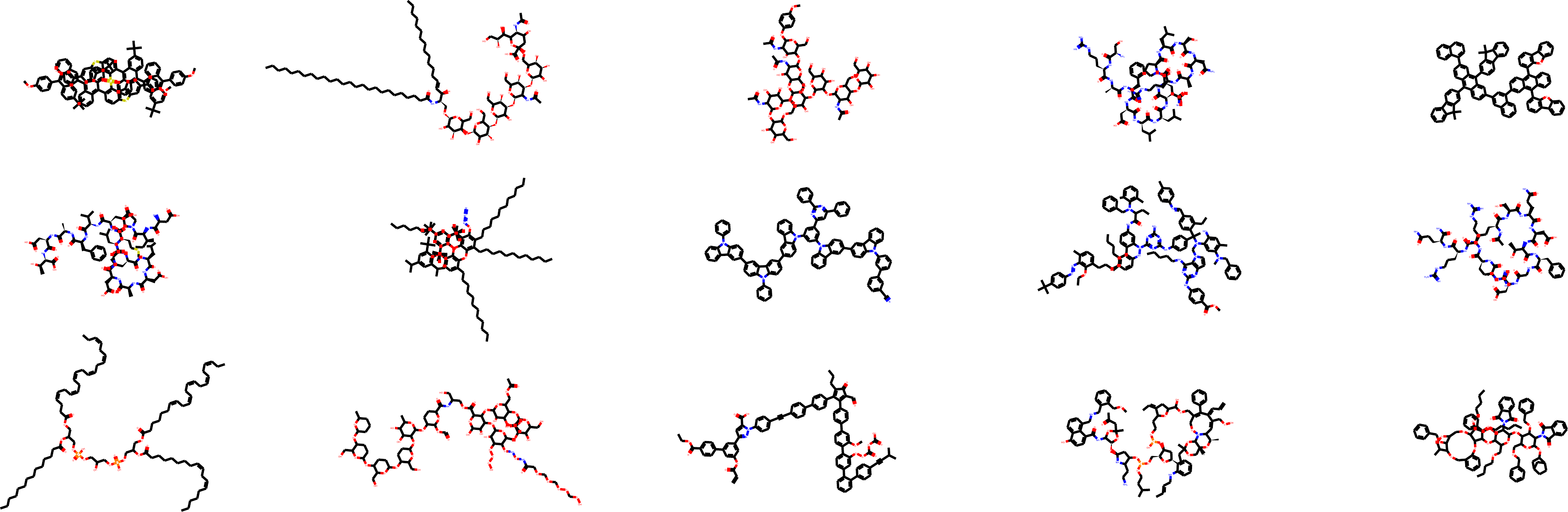}};
    \draw (-8.85, 2.75) node[] {\textbf{b}};
    \draw (-8.85, 1.95) node[rotate=90] {TRAIN};
    \draw (-8.85, 0.2)  node[rotate=90] {SM-RNN};
    \draw (-8.85, -1.80) node[rotate=90] {SF-RNN};
\end{tikzpicture}

\vspace{0.25cm}

\begin{tikzpicture}
    \draw (0, 0) node[inner sep=0] {\includegraphics[width=0.98\textwidth]{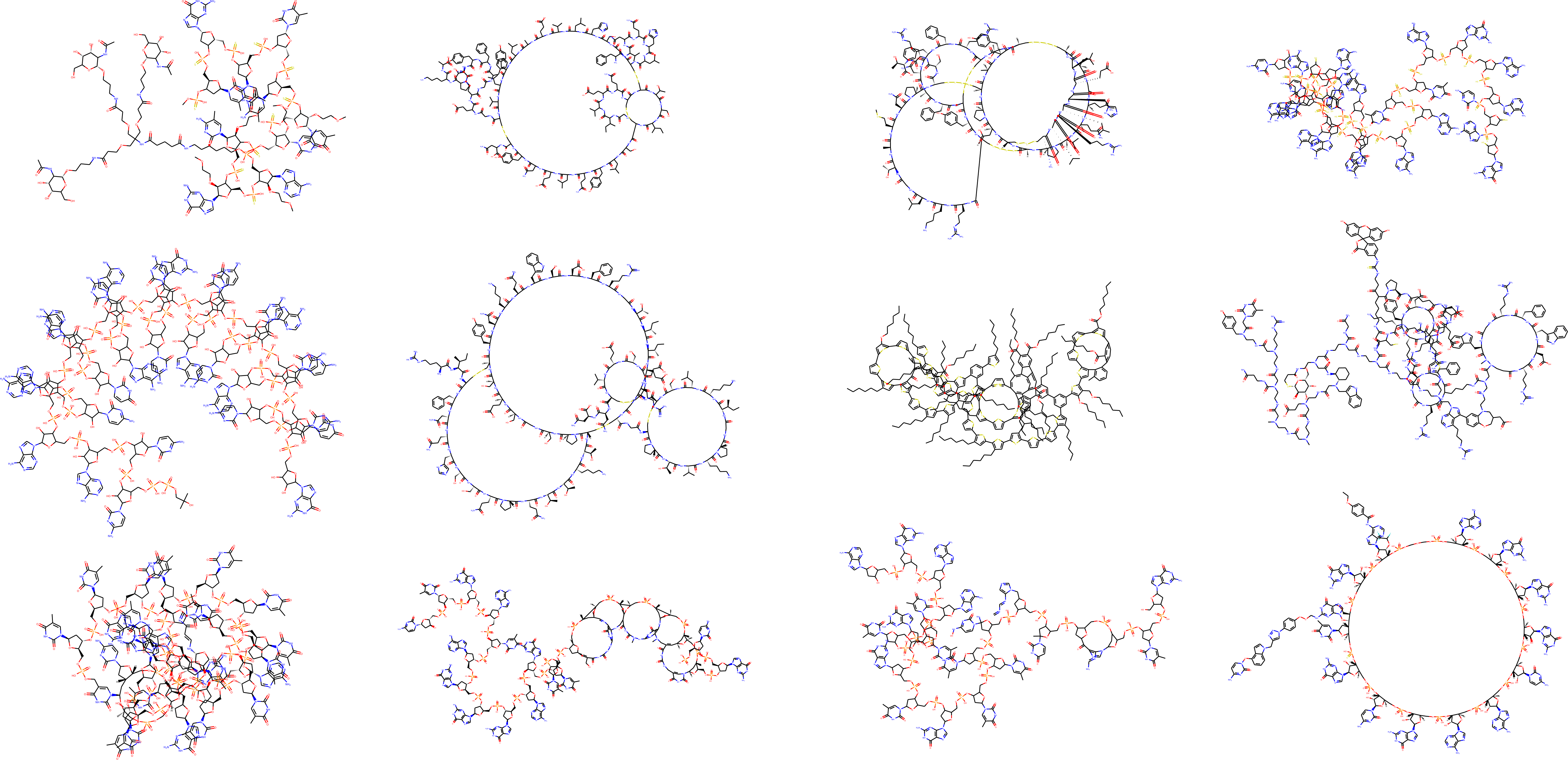}};
    \draw (-8.85, 4.10) node[] {\textbf{c}};
    \draw (-8.85, -2.70) node[rotate=90] {SF-RNN};
    \draw (-8.85, 0.68) node[rotate=90] {SM-RNN};
     \draw (-8.85, 3.05) node[rotate=90] {TRAIN};
\end{tikzpicture}

\caption{\textbf{Large Scale Task}   
\ \textbf{a} Histograms of fragment \#, single atom fragment \#, double atom fragment \#, single ring fragment \#, multi-ring  fragment \#   
(all per molecule) from molecules generated by the RNN models or from the training data 
\ \textbf{b} Five molecules generated by the RNN models or from the training data with $ 1500 \leq $ Molecular Weight $\leq 2000$.
\ \textbf{c} Four molecules generated by the RNN models or from the training data with Molecular Weight $\geq 4000$}
\label{fig:ls2}
\end{figure*}

\section{Results}

We define three tasks, generating: 1) distributions of molecules with high scores of penalized LogP \cite{gomez2018automatic} (FIG. \ref{fig:tasks} \textbf{a,d}) 
2) multi-modal distributions of molecules (FIG. \ref{fig:tasks} \textbf{b,e}) and 
3) the largest molecules in PubChem  (FIG. \ref{fig:tasks} \textbf{c,e}). 
Necessarily, each different generative modeling task is defined by learning to generate from the distribution of molecules in a dataset. 
We build three datasets using relevant subsets of larger databases.

For each task we assess performance by plotting the distribution of training molecules properties and the distribution learned by the language models and graph models. 
We use a histogram for the training molecules and fit a Gaussian kernel density estimator to it by tuning its bandwidth parameter. 
We plot KDE's for molecular properties from all models using the same bandwidth parameter. 

From all models we initially generate 10K (thousand) molecules, compute their properties and use them to produce all plots and metrics. 
Furthermore, for fair comparison of learned distributions, we use the same number of generated molecules from all models after removing duplicates and training molecules.

For quantitative evaluation of any model's ability to learn its training distribution, we compute the Wasserstein distance between property values of generated molecules and training molecules. 
We also compute the Wasserstein distance between different samples of training molecules in order to determine a most optimal baseline, which we can compare with as an oracle.  

For molecular properties we consider: 
quantitative estimate of drug-likeness (QED) \cite{bickerton2012quantifying}, 
synthetic accessibility score (SA) \cite{ertl2009estimation},  
octanol-water partition coefficient (LogP) \cite{wildman1999prediction}, 
exact molecular weight (MW), Bertz complexity (BCT) \cite{bertz1981first}, 
natural product likeness (NP) \cite{ertl2008natural}. We also use standard metrics like validity, uniqueness, novelty-- to assess the model's ability to generate a diverse set of real molecules distinct from the training data.  

For models, our main consideration is a chemical language model using a recurrent neural network with long short-term memory \cite{hochreiter1997long} and is trained on SMILES (SM-RNN) or SELFIES (SF-RNN). 
We also train two state of the art graph generative models: the junction tree variational autoencoder (JTVAE) \cite{jin2018junction} and the constrained graph variational autoencoder (CGVAE) \cite{liu2018constrained}. 
Each model is trained on a different molecular representation: SMILES/SELFIES strings or trees/graphs (FIG. \ref{fig:tasks} \textbf{g}).

\begin{table}[t]
\begin{tabular}{c| c cccccc}
               & LogP  &  SA   & QED  & MW & BCT & NP \\ 
         \hline \noalign{\smallskip}
         TRAIN  & 0.020 & 0.0096  & 0.0029  & 1.620 & 7.828 &  0.013 \\   
         \hline \noalign{\smallskip}
         SM-RNN & 0.095 & 0.0312  & 0.0068  & 3.314 & 21.12 &  0.054   \\
         SF-RNN & 0.177 & 0.2903  & 0.0095  & 6.260 & 25.00 &  0.209  \\
         JTVAE & 0.536 & 0.2886  & 0.0811  & 35.93 & 76.81 &  0.164  \\
         CGVAE & 1.000 & 2.1201  & 0.1147  & 69.26 & 141.2 &  1.965  \\
\end{tabular} \caption{\textbf{Penalized LogP Task} \ Wasserstein metrics for LogP, SA, QED, MW, BT and NP between molecules from the training data and generated by the models. TRAIN is an oracle baseline- values closer to it are better.}
\label{tab:logp}
\end{table}

\subsection{Penalized LogP Task} 

For the first task, we consider one of the most widely used benchmark assessments for searching chemical space, the penalized LogP task-- finding molecules with high LogP \cite{ghose1986atomic} penalized by synthesizability \cite{ertl2009estimation} and unrealistic rings. 
We turn this task into a generative modeling one, where the goal is to learn distributions of molecules with high penalized LogP scores. 
Finding a single molecule with a good score (above 3.0) can be challenging but learning to directly generate from this part of chemical space, so that every molecule produced by the model has high penalized LogP, adds another layer of complexity to the standard task. For this we build a training dataset by screening the ZINC15 database \cite{irwin2005zinc} for molecules with values of penalized LogP exceeding 4.0. Most machine learning approaches only find a handful of molecules in this range, for example JTVAE \cite{jin2018junction} found 22 total during all their attempts. After screening, the top scoring molecules in ZINC amounted to roughly ~160K (K is thousand) molecules for the training data in this task. Thus, the training distribution is extremely spiked with most density falling around 4.0-4.5 penalized LogP as seen in FIG.\ref{fig:tasks}\textbf{a} with most training molecules resembling the examples shown in FIG. \ref{fig:tasks} \textbf{d}. However, some of the training molecules, around 10\% have even higher penalized LogP scores-- adding a subtle tail to the distribution.

The results of training all models are shown in FIG. \ref{fig:logp} and \ref{fig:hlogp}. 
The language models perform significantly better than the graph models, 
with the SELFIES RNN producing a slightly closer match to the training distribution in FIG. \ref{fig:logp}\textbf{a}. The CGVAE and JTVAE learn to produce a large number of molecules with penalized LogP scores that are substantially worse than the lowest training scores. It is important to note, from the examples of these shown in FIG. \ref{fig:logp}\textbf{a} these lower scoring molecules are quite similar to the molecules from the main mode of the training distribution, this highlights the difficulty of learning this distribution. In FIG. \ref{fig:logp}\textbf{b} we see that JTVAE and CGVAE learn to produce more molecules with larger SA scores than the training data. For LogP we see that all models learn the main mode but the RNNs produce closer distributions, similar results can be seen for QED. These results carryover for quantitative metrics and both RNNs achieve lower Wasserstein distance metrics than the CGVAE and JTVAE (Table \ref{tab:logp}) with the SMILES RNN coming closest to the TRAIN oracle.  

We further investigate the highest penalized LogP region of the training data with values exceeding 6.0-- the subtle tail of the training distribution. In the 2d distributions (FIG. \ref{fig:hlogp}\textbf{a}) it's clear that
both RNNs learn this subtle aspect of the training data while the graph models ignore it almost completely and only learn molecules that are closer to the main mode. 
In particular, CGVAE learns molecules with larger SA score than the training data. Furthermore, the molecules with highest penalized LogP scores in the training data typically contain very long carbon chains and fewer rings (FIG. \ref{fig:hlogp}\textbf{b})-- the RNNs are capable of picking up on this. This is very apparent in samples shown from the baselines and training data in FIG. \ref{fig:hlogp}\textbf{c}-- the RNNs produce similar molecules while the CGVAE and JTVAE generate molecules with many rings that have penalized LogP scores near 6.0. The language models produce molecules more similar to the training examples in  
 FIG. \ref{fig:hlogp}\textbf{c} and their distribution is close to the training distribution in the histograms of FIG. \ref{fig:hlogp}\textbf{a}\& \textbf{b}. 
Overall, the language models could learn distributions of molecules with high penalized LogP scores, better than the graph models.

\begin{table}[t]
\begin{tabular}{c|ccccccc}
       & LogP  &  SA    & QED & MW & BCT & NP \\
      \hline \noalign{\smallskip}
        TRAIN & 0.048 & 0.0158  & 0.0020  & 2.177 & 14.149 & 0.010  \\ 
       \hline \noalign{\smallskip}
         SM-RNN & 0.081 & 0.0246  & 0.0059  & 5.483 & 21.118 & 0.012  \\
         SF-RNN & 0.286 & 0.1791  & 0.0227  & 11.35 & 68.809 & 0.079  \\
         JTVAE & 0.495 & 0.2737  & 0.0343  & 27.71 & 171.87 & 0.109  \\
         CGVAE & 1.617 & 1.8019  & 0.0764  & 30.31 & 183.58 & 1.376  \\
\end{tabular} \caption{\textbf{Multi-distribution Task} \ Wasserstein metrics of LogP, SA, QED, MW, BCT, NP 
of training and model generated molecules. TRAIN is an oracle baseline.  }
\label{tab:md}
\end{table}

\subsection{Multi-distribution Task} 

For the next task, we created a dataset by combining subsets of : 1) GDB13 \cite{blum2009970} molecules with molecular weight (MW) $\leq$ 185 2) 
ZINC \cite{gomez2018automatic,irwin2005zinc} molecules with 185 $\leq$  MW $\leq$ 425
3) Harvard clean energy project (CEP) \cite{hachmann2011harvard} molecules with 460 $\leq$  MW $\leq$ 600, and the
4) POLYMERS \cite{st2019message} molecules with MW $\geq$ 600.
The training distribution has four modes-- (FIG.\ref{fig:tasks}\textbf{b},\textbf{e} \& \ref{fig:md}\textbf{a}). 
CEP \& GDB13 make up 1/3 and ZINC \& POLYMERS take up 1/3 each of $\sim$ 200K training molecules.  

In the multi-distribution task, both RNN models capture the data distribution quite well and learn every mode in the training distribution FIG. \ref{fig:md}\textbf{a}, with comparable quality. On the other hand, JTVAE entirely misses the first mode from GDB13 then poorly learns ZINC and CEP. As well, CGVAE learns GDB13 but underestimates ZINC and entirely misses the mode from CEP. More evidence that the RNN models learn the training distribution more closely is apparent in FIG. \ref{fig:md}\textbf{c} where CGVAE and JTVAE barely distinguish the main modes. Additionally, the RNN models generate molecules better resembling the training data (supplementary FIG. \ref{fig:mds}). Despite this, all models-- except CGVAE, capture the training distribution of QED, SA score and Bertz Complexity (FIG. \ref{fig:md}\textbf{b}). 
Lastly, in TABLE \ref{tab:md} the RNN trained on SMILES has the lowest Wasserstein metrics followed by the SELFIES RNN then JTVAE and CGVAE.

\subsection{Large Scale Task} 

The last generative modeling task, involves testing the ability of deep generative models to learn large molecules, the largest possible molecules relevant to molecular generative models that use SMILES/SELFIES string representations or graphs. For this we turn to PubChem \cite{kim2016pubchem} and screen for the largest molecules with more than 100 heavy atoms, producing ~300K molecules. These are molecules of various kinds: small bio-molecules, photovoltaics and others (FIG.\ref{fig:tasks}\textbf{f}). 
They also have a wide range of molecular weight from 1500 to 5000 but most molecules fall into the 1500-2000 range (FIG.\ref{fig:tasks}\textbf{c}).  

This task was the most challenging for the graph models, both failed to train and were entirely incapable of learning the training data. 
In particular, JTVAE's tree decomposition algorithm applied to the training data produced a fixed vocabulary of $\sim$11,000 substructures. 
However both RNN models were able to learn to generate molecules as large and as varied as the training data. 
The training molecules correspond to very long SMILES and SELFIES string representations, in this case, the SELFIES strings provided an additional advantage and the SELFIES RNN could match the data distribution more closely (FIG. \ref{fig:ls}\textbf{a}). 
In particular, learning valid molecules is substantially more difficult with the SMILES grammar, as there are many more characters to generate for these molecules and
a higher probability that the model will make a mistake and produce an invalid string. In contrast, the SELFIES string generated will never be invalid. 
Interestingly, even when the RNN models generated molecules that were out of distribution and substantially smaller than the training molecules-- they still had similar substructures and resemblance to the training molecules (\ref{fig:ls}\textbf{a}). In addition, the training molecules seemed to be divided into two modes of molecules with lower and higher LogP values (FIG. \ref{fig:ls}\textbf{b}): with biomolecules defining the lower mode and molecules with more rings and longer carbons chains defining the higher LogP mode (more example molecules can be seen in supplementary FIG. \ref{fig:largelogp}). The RNN models were both able to learn the bi-modal nature of the training distribution.   

The training data has a variety of different molecules and substructures, in FIG. \ref{fig:ls2}\textbf{a} the RNN models adequately learn the distribution of substructures arising in the training molecules. Specifically the distribution for the number of: fragments, single and double atom fragments as well as single and multi-rings fragments in each molecule. As the training molecules get larger and occur less, both RNN models still learn to generate these molecules (FIG. \ref{fig:ls}\textbf{a} when molecular weight$>$3000). To demonstrate the size of the molecules being generated a few examples from both RNN models and training data are shown, from the main mode of the distribution (FIG. \ref{fig:ls2}\textbf{b}) and largest training molecules with weight larger than 4000 (FIG. \ref{fig:ls2}\textbf{c}).

\begin{table}[t]
  \begin{tabular}{@{}c|l|cccc@{}}
    Task &  Metric      & SM-RNN & SF-RNN & JTVAE  & CGVAE   \\
 \hline \noalign{\smallskip}
  \parbox[t]{4mm}{\multirow{3}{*}{\rotatebox[origin=c]{90}{LogP}}}
     & validity     & 0.941    & 1.000 & 1.000 & 1.000  \\
     & uniqueness   & 0.987    & 1.000 & 0.982 & 1.000  \\    
     & novelty      & 0.721    & 0.871 & 0.980 & 1.000  \\
 \hline \noalign{\smallskip}
  \parbox[t]{4mm}{\multirow{3}{*}{\rotatebox[origin=c]{90}{Multi}}}
     & valid          & 0.969    & 1.000	& 0.999   &   0.999   \\
     & uniqueness     & 0.996    & 0.989	& 0.998   & 0.996  \\    
     & novelty        & 0.937    & 0.950	& 0.998   & 1.000  \\
  \hline \noalign{\smallskip}
  \parbox[t]{4mm}{\multirow{3}{*}{\rotatebox[origin=c]{90}{Large}}}
     & valid          & 0.876    &  1.000 &  -      & -      \\
     & uniqueness     & 0.999    &  0.994 &  -      & -      \\    
     & novelty        & 0.999    &  0.999 &  -      & -      \\
  \end{tabular}
 \caption{\textbf{Standard Metrics} \ Validity, uniqueness and novelty of molecules generated by all models in every task. Closer to 1.0 indicates better performance.}
\label{tab:metrics}
\end{table}

\subsection{Metrics}

We also evaluate models on standard metrics in the literature: validity, uniqueness and novelty.  
Using the same 10K molecules generated from each model for each task we compute the following statistics defined in \cite{simonovsky2018graphvae} and store them in TABLE. \ref{tab:metrics}: 
1) validity: the ratio between the number of valid and generated molecules, 
2) uniqueness: the ratio between the number of unique molecules (that are not duplicates) and valid molecules,
3) novelty: the ratio between unique molecules that are not in the training data and the total number of unique molecules. 
In the first two tasks (TABLE. \ref{tab:metrics}), JTVAE and CGVAE have better metrics with very high validity, uniqueness and novelty (all close to 1), 
here the SMILES and SELFIES RNN perform worse but the SELFIES RNN is close to their performance. 
The SMILES RNN has the worse metrics due to its poor grammar but is not substantially worse than the other models.

\section{Discussion}

In this work, in effort to test the ability of chemical language models, 
we introduce three complex modeling tasks for deep generative models of molecules. 
Language models and graph baselines perform each task, which entails learning to generate molecules from a challenging dataset to learn.
The results demonstrate that language models are very powerful, flexible models that can learn a variety of very different complex distributions while the popular graph baseline are much less capable. 

\noindent \textbf{SELFIES improves performance, decreases memorization.} \quad
While the SMILES RNN still achieves impressive performance in all tasks, we notice that the SELFIES representation seems to improve the performance of language models in every task, 
particularly in the large scale task 
and consistently yields better standard metrics of validity, uniqueness and novelty. 
Additionally, the SELFIES grammar may be more difficult to overfit to, which is the likely explanation for why the SELFIES RNN has better standard metrics but worse Wasserstein metrics than the SMILES RNN. 
The SMILES RNN with lower novelty scores and Wasserstein metrics that are very close to the TRAIN oracle may be memorizing more of the training data than the SELFIES RNN.
In future, it would be valuable to ascertain if one should ever use SMILES over SELFIES. 

\noindent \textbf{Graph generative models are not flexible} \quad
The results show that the most popular graph generative models: 
JTVAE and CGVAE had trouble learning the distributions in each task.
For the penalized LogP task,
the difference between a molecule that has a score of 2 and one that scores 4 often can be very subtle. 
Sometimes changing a single carbon or other atom can cause a large drop in score-- 
this likely explains why the CGVAE severely misfit the main training mode. 
For the multi-distribution task, JTVAE and CGVAE's difficulties are clear but very understandable. For JTVAE, it has to learn a wide range of tree types: many of which have no large substructures like rings (the GDB13 molecules) while others are entirely rings (some of the CEP and POLYMERS). For CGVAE, it has to learn a wide range of very different generation traces-- which is difficult especially since it only uses one sample trace during learning. For the same reasons, these models were incapable of training on the largest molecules in PubChem.

Based on the experiments conducted, language models are very powerful generative models for learning any complex molecular distribution and should see even more widespread use. 
However, it is still possible to see improvements to these models, especially in combating over-fitting as the results 
did give evidence that language models can suffer from this. 
In addition, we hope that the molecular modeling tasks and datasets introduced can motivate new generative models that address the difficulties that the graph baselines experienced.  
Future work will explore how capable chemical language models are in learning larger and larger snapshots of chemical space.

\section{Methods}

\noindent \textbf{Model details and hyper-parameters:} For each task we give the model and training hyper-parameters of the best model found during hyper-parameter optimization, where for all models we used random search with 100 randomly sampled hyper-parameters sets. 

\noindent \textbf{Penalized LogP Task.} For the SM-RNN we used an LSTM with 2 hidden layer with 400 units and dropout in the last layer with prob=0.2 and learning rate of 0.0001. 
For the  SF-RNN we used an LSTM with 2 hidden layer with 600 units and dropout in the last layer with prob=0.4 and learning rate of 0.0002. 
The CGVAE used  8 propagation layers and hidden layer side of 100 with kl annealed to 0.1 and a learning rate of 0.0015.
The JTVAE used a learning rate of 0.001 and 3 GNN layers with a hidden size of 356.

\textbf{Multi-Distribution Task.}
For the SM-RNN we used an LSTM with 3 hidden layer with 512 units and dropout in the last layer with prob=0.5 and learning rate of 0.0001. 
For the  SF-RNN we used an LSTM with 2 hidden layer with 500 units and dropout in the last layer with prob=0.2 and learning rate of 0.0003. 
The CGVAE used  8 propagation layers and hidden layer side of 100 with kl annealed to 0.1 and a learning rate of 0.001.
The JTVAE used a learning rate of 0.0001 and 3 GNN layers with a hidden size of 356.

\textbf{Large-Scale Task.}
For the SM-RNN we used an LSTM with 2 hidden layer with 512 units and dropout in the last layer with prob=0.25 and learning rate of 0.001. 
For the  SF-RNN we used an LSTM with 2 hidden layer with 800 units and dropout in the last layer with prob=0.4 and learning rate of 0.0001. 

\section{Data Availability}
Training data and generated molecules is available: 
\href{https://github.com/danielflamshep/genmoltasks}{https://github.com/danielflamshep/genmoltasks}.

\section{Code Availability}

The code used to train models is from public repos. 

\noindent JTVAE: 
\href{https://github.com/wengong-jin/icml18-jtnn}{https://github.com/wengong-jin/icml18-jtnn}.

\noindent CGVAE: 
\href{https://github.com/microsoft/constrained-graph-variational-autoencoder}{https://github.com/microsoft/constrained-graph-variational-autoencoder}. 

The RNN models were trained using the char-rnn code from 
\href{https://github.com/molecularsets/moses}{https://github.com/molecularsets/moses}. 
Trained models are available upon request.

\section{Acknowledgements}   A.A.-G. acknowledge funding from Dr. Anders G. Fr{\o}seth.
A.A.-G. also acknowledges support from the Canada 150 Research Chairs Program, the Canada Industrial Research Chair Program, and from Google, Inc.
Models were trained using the Canada Computing Systems \cite{baldwin2012compute}.

\bibliographystyle{apsrev4-1}
\bibliography{main}

\pagebreak
\begin{widetext}

\newpage

\section*{Supplemental Materials}

\setcounter{equation}{0}
\setcounter{figure}{0}
\setcounter{table}{0}
\makeatletter
\renewcommand{\theequation}{S\arabic{equation}}
\renewcommand{\thefigure}{S\arabic{figure}}

\begin{figure}[h]
    \centering
    \includegraphics[width=0.99\textwidth]{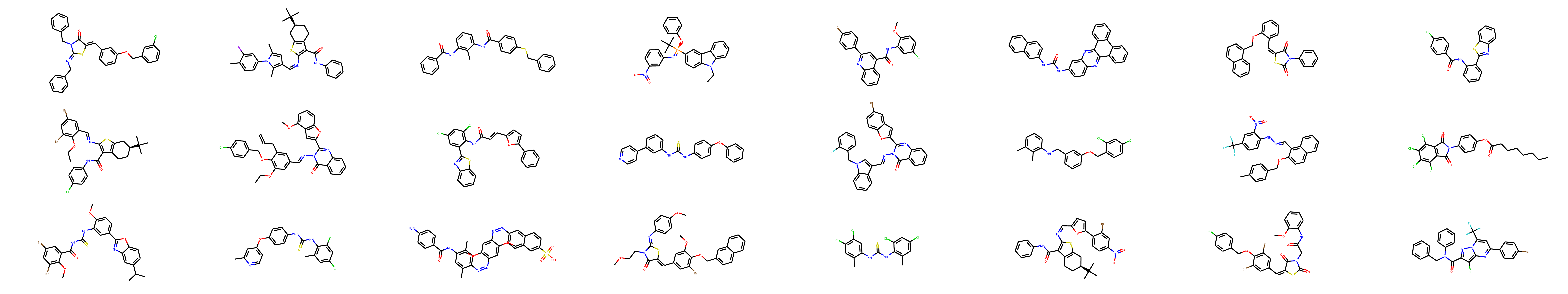}

    \textbf{a } Molecules with  $4 \leq \text{ penalized LogP} \leq 5$.
    
    \vspace{0.125cm}

    \includegraphics[width=0.99\textwidth]{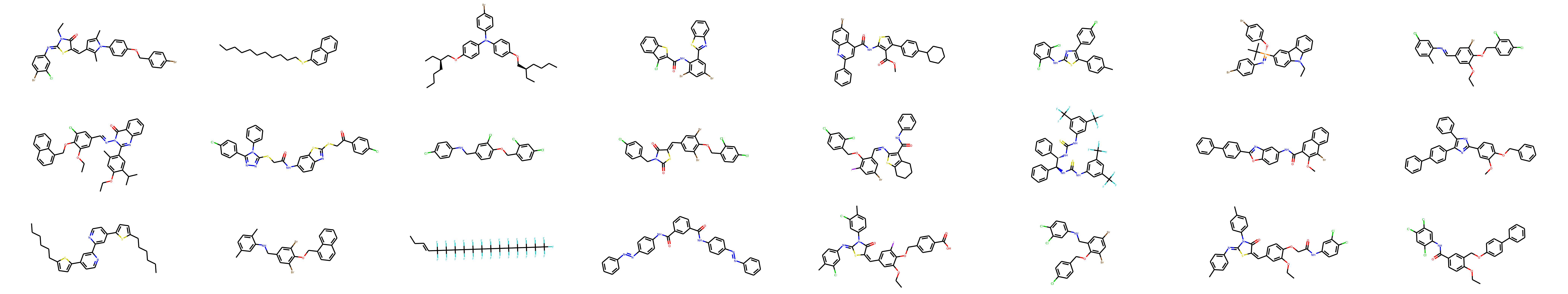}

    \textbf{b } Molecules with  $5 \leq \text{ penalized LogP} \leq 6$.

    \vspace{0.125cm}

    \includegraphics[width=0.99\textwidth]{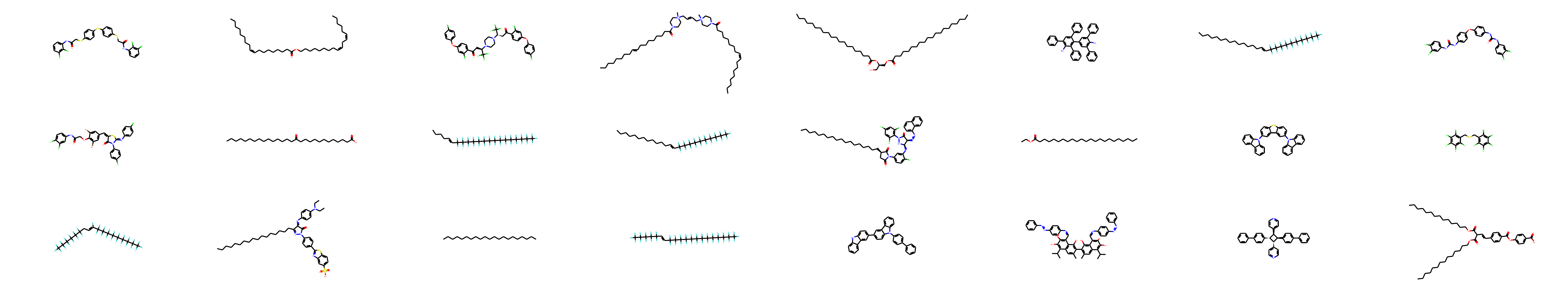}

    \textbf{c } Molecules with  $6 \leq \text{ penalized LogP} \leq 8$.
    
    \vspace{0.125cm}

    \includegraphics[width=0.99\textwidth]{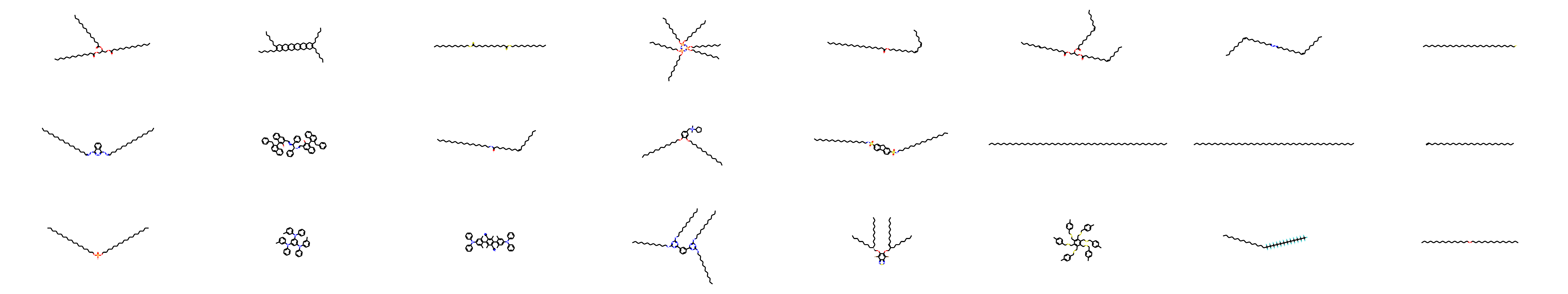}

    \textbf{d } Molecules with  $\text{ penalized LogP} \geq 8$.
    
    \vspace{0.125cm}

    \includegraphics[width=0.99\textwidth]{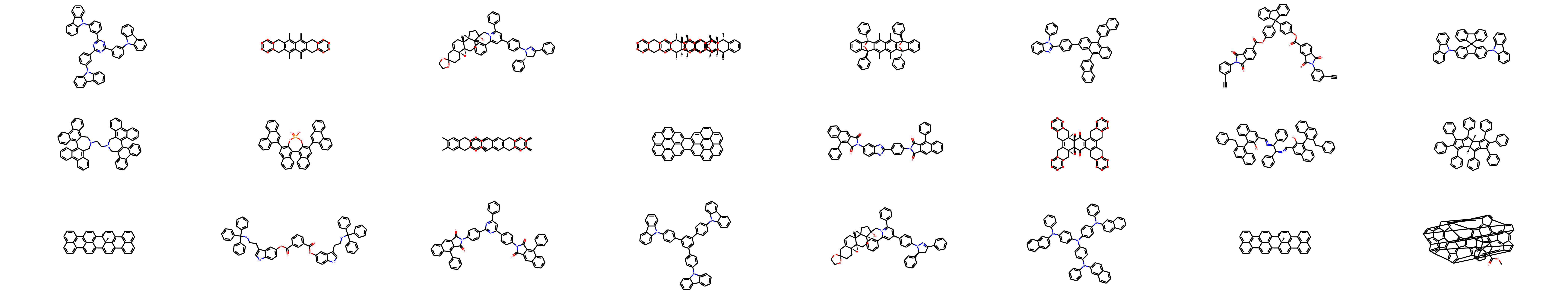}

    \textbf{e } Molecules with more than 10 rings.
    
    \vspace{0.125cm}

    \caption{\textbf{Penalized LogP Task} \textbf{a-e} Training molecules with different properties.}
    \label{fig:logptrain}
\end{figure}

\begin{table*}[t] \raggedleft
\begin{tabular}{l|c}
& Generated Molecules \\ \hline \hline
\rotatebox{90}{\hspace{1.25cm} \textbf{TRAIN}}  & \includegraphics[width=0.99\textwidth]{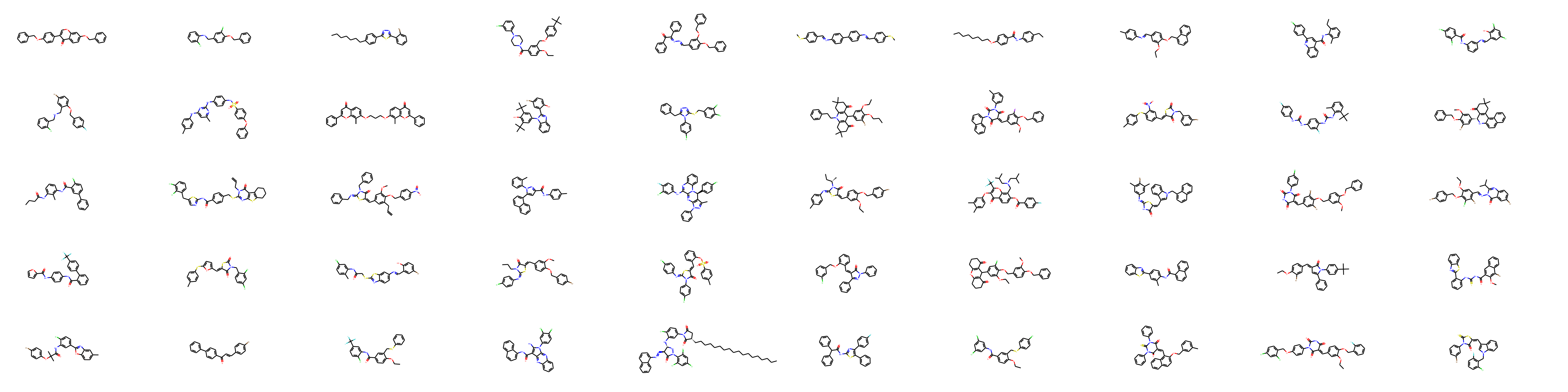} \\
\noalign{\smallskip} \hline    \noalign{\smallskip}  \vspace{-0.25cm} \\ 
\rotatebox{90}{\hspace{1.5cm} \textbf{CGVAE}}  & \includegraphics[width=0.99\textwidth]{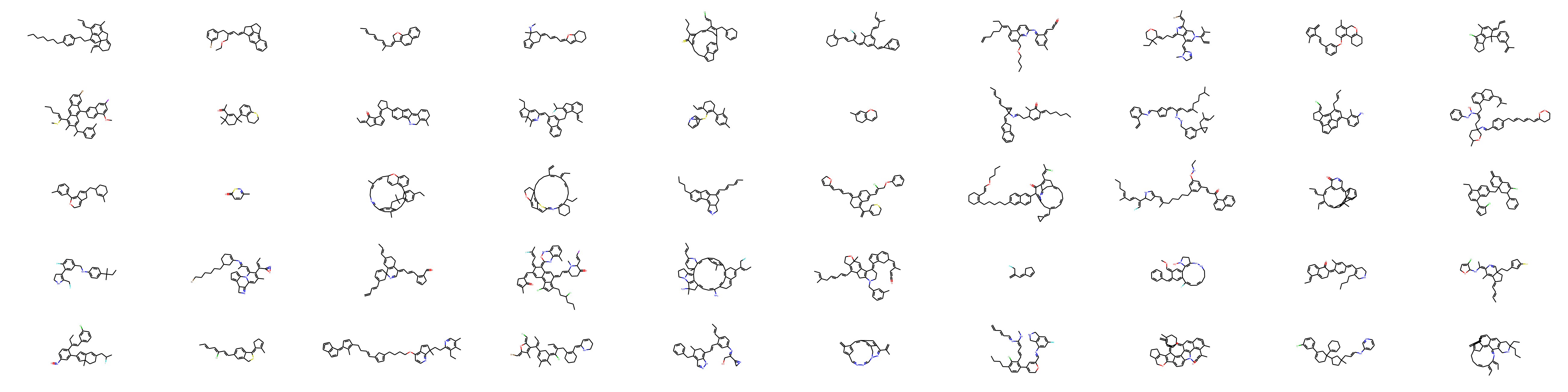} \\
\noalign{\smallskip} \hline    \noalign{\smallskip}  \vspace{-0.25cm} \\ 
\rotatebox{90}{\hspace{1.5cm} \textbf{JTVAE}}  & \includegraphics[width=0.99\textwidth]{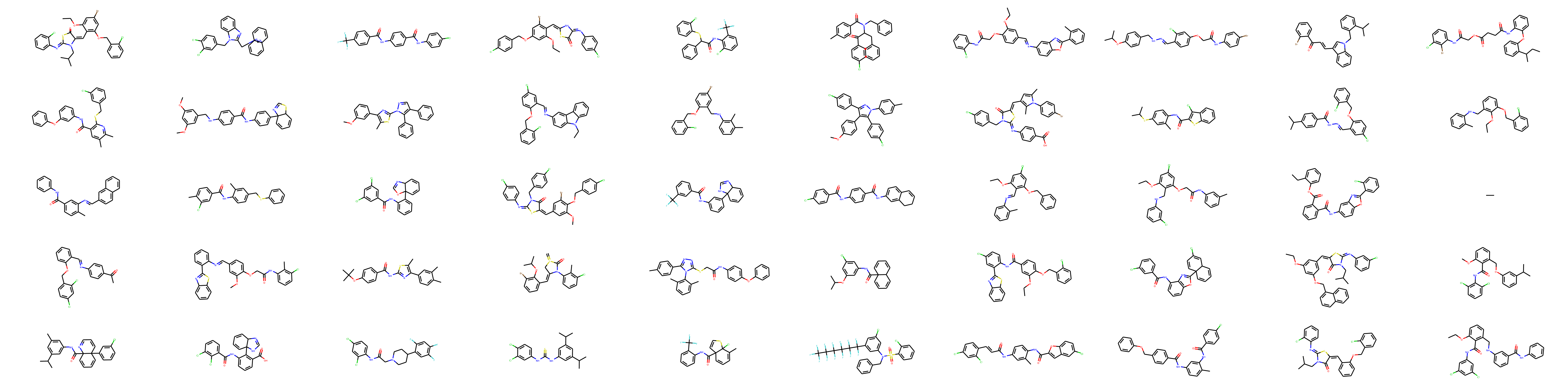} \\
\noalign{\smallskip} \hline    \noalign{\smallskip}  \vspace{-0.25cm} \\ 
\rotatebox{90}{\hspace{1.5cm} \textbf{SF-RNN}}  & \includegraphics[width=0.99\textwidth]{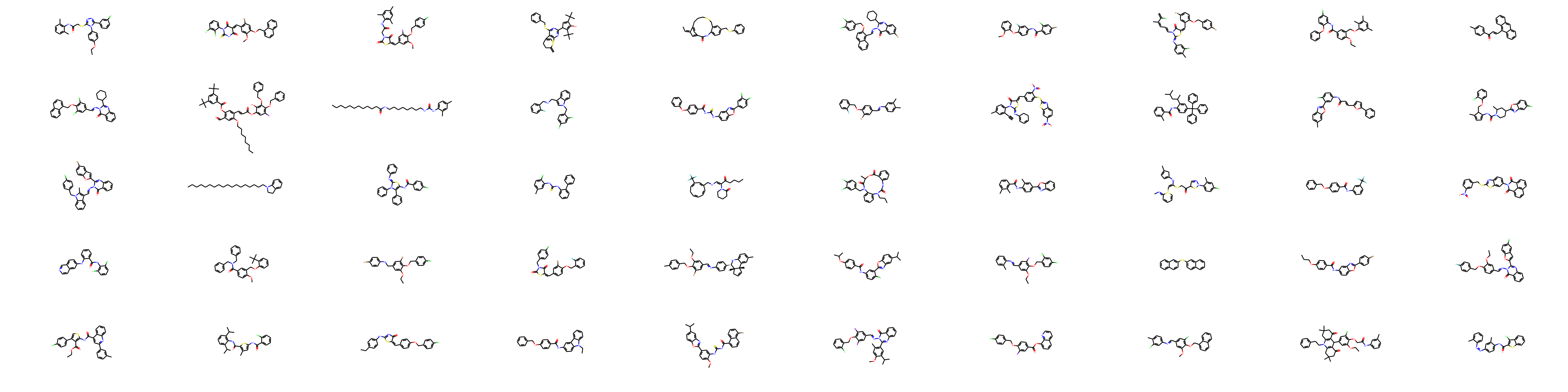} \\
\noalign{\smallskip} \hline    \noalign{\smallskip}  \vspace{-0.25cm} \\ 
\rotatebox{90}{\hspace{1.5cm} \textbf{SM-RNN}}  & \includegraphics[width=0.99\textwidth]{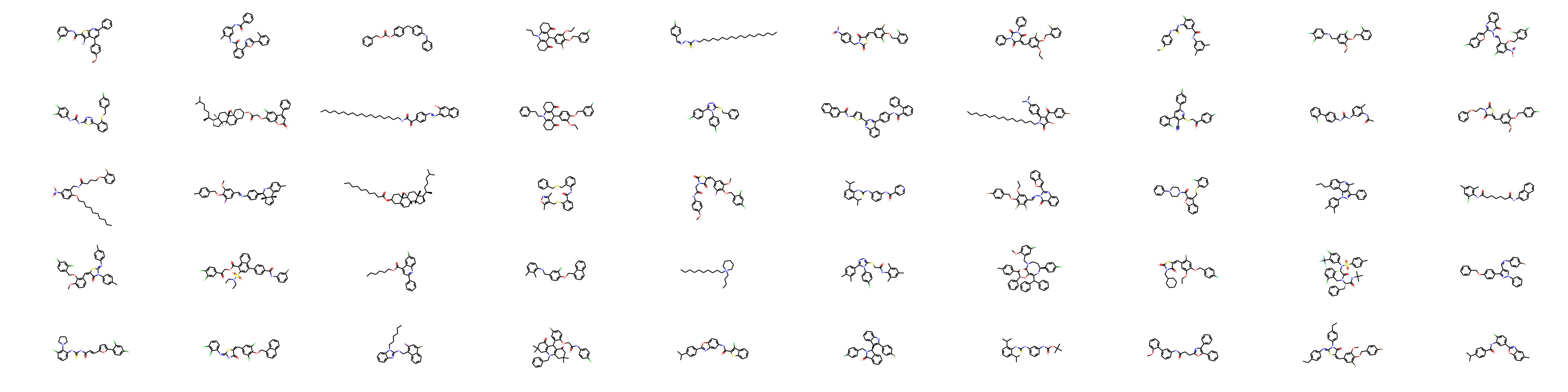} 
\end{tabular}
\caption{\textbf{Penalized LogP Task } Molecules generated from each model.}
\label{fig:logpmodels}
\end{table*}

\begin{figure}[t]
    \centering
    \includegraphics[width=0.925\textwidth]{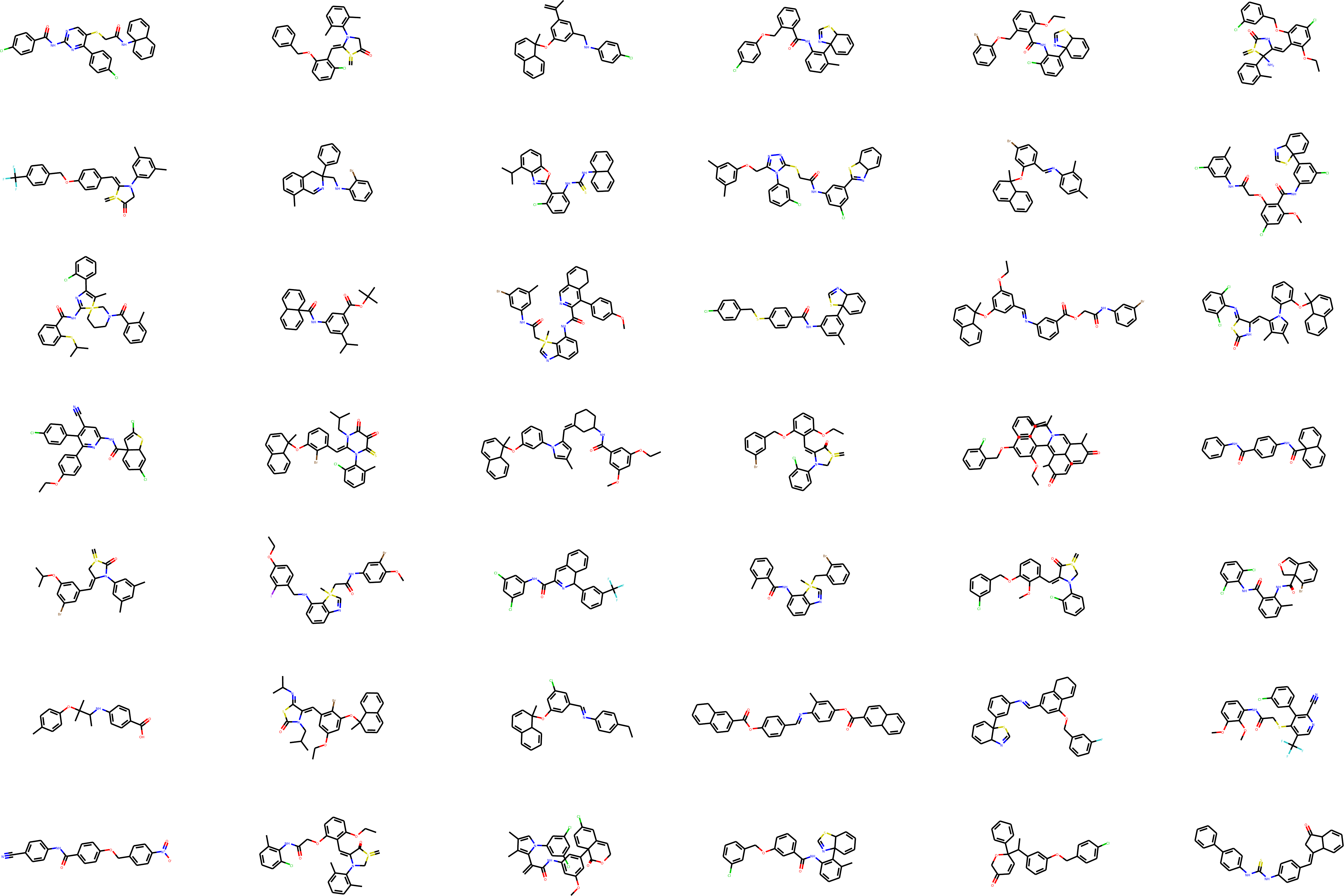}
    
    \vspace{0.25cm}

    \textbf{a } Molecules from JTVAE. 
    
     \includegraphics[width=0.925\textwidth]{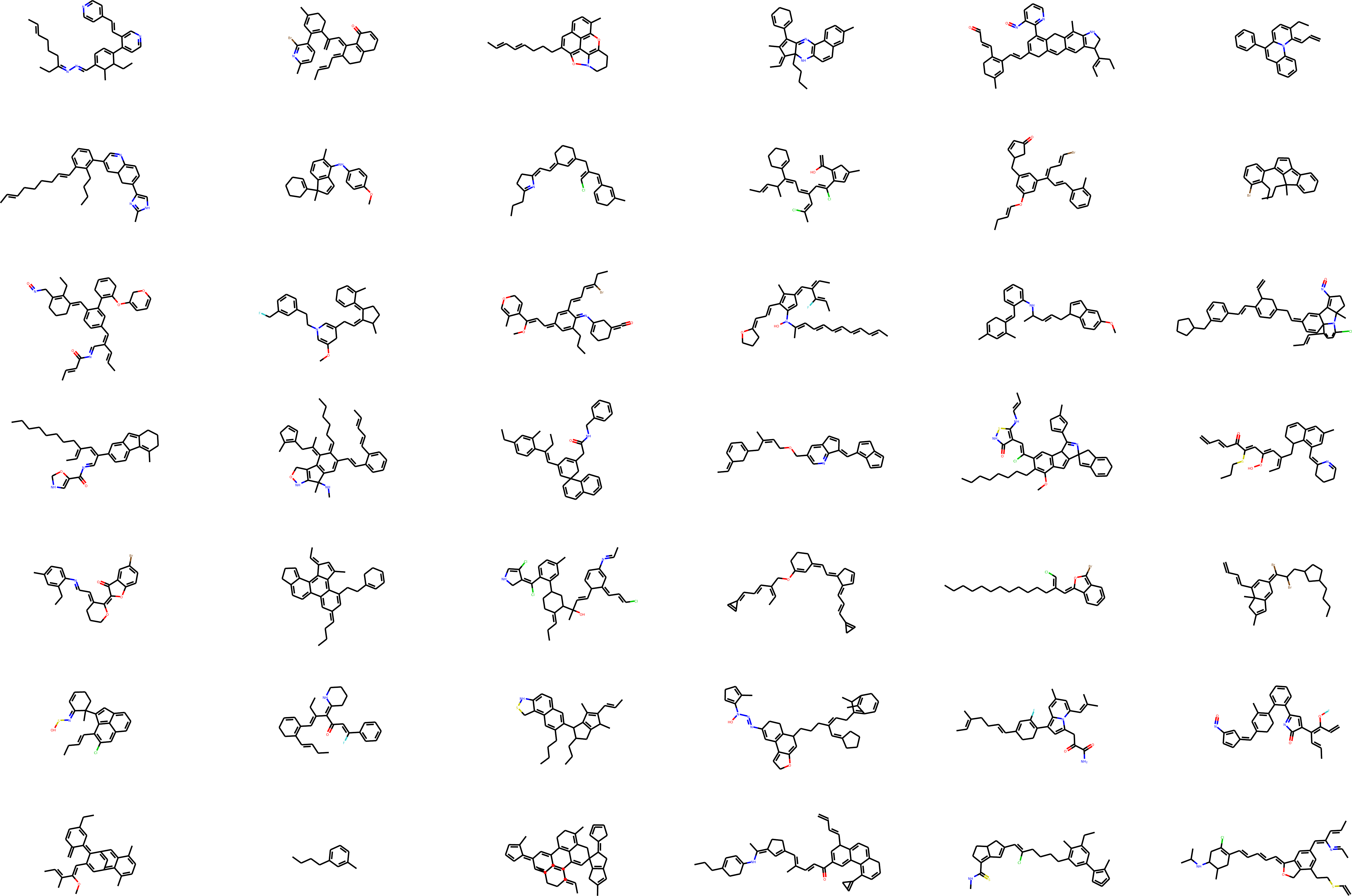}
    
    \vspace{0.25cm}

    \textbf{b } Molecules from CGVAE. 
    
    \caption{\textbf{Penalized LogP Task} \textbf{a-b} Generated molecules with penalized LogP $<4$ from the graph generative models.}
    \label{fig:logpood}
\end{figure}

\begin{table*}[t] \raggedleft
\begin{tabular}{l|c}
& Generated molecules \\ \hline \hline
\rotatebox{90}{\hspace{1cm} \textbf{TRAIN}}  & \includegraphics[width=0.99\textwidth]{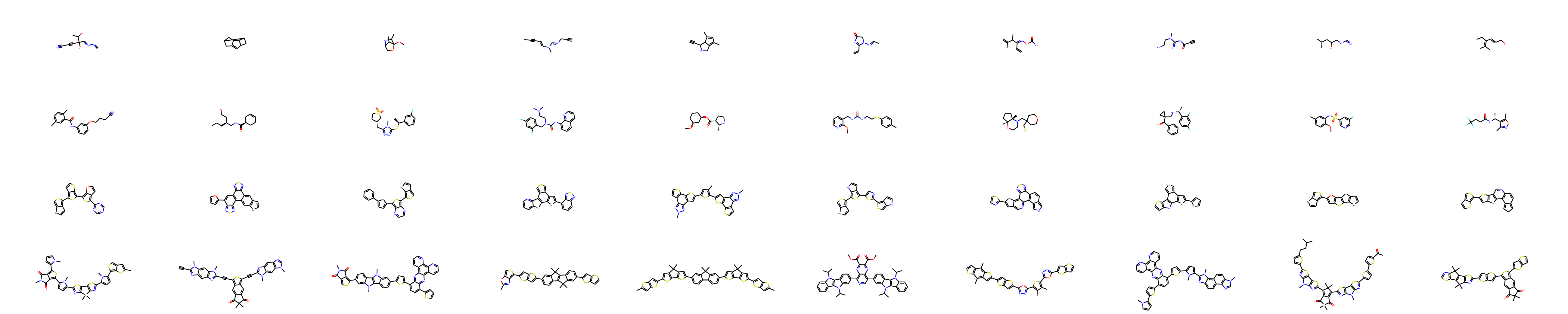} \\
\noalign{\smallskip} \hline    \noalign{\smallskip}  \vspace{-0.25cm} \\ 
\rotatebox{90}{\hspace{1cm} \textbf{CGVAE}}  & \includegraphics[width=0.99\textwidth]{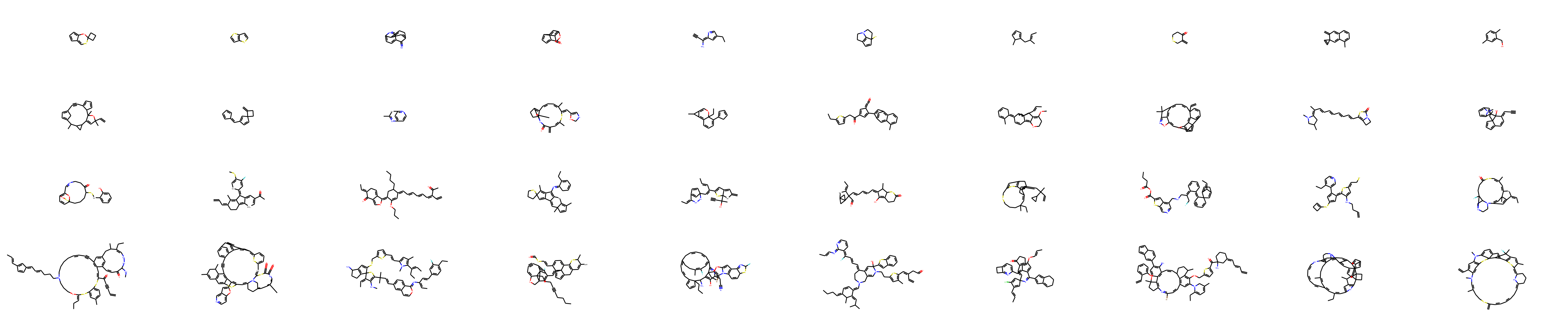} \\
\noalign{\smallskip} \hline    \noalign{\smallskip}  \vspace{-0.25cm} \\ 
\rotatebox{90}{\hspace{1cm} \textbf{JTVAE}}  & \includegraphics[width=0.99\textwidth]{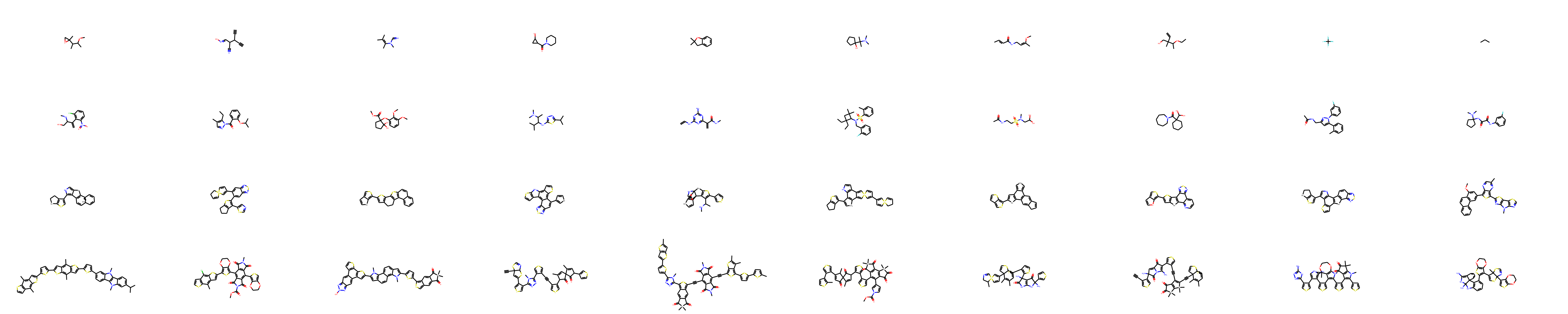} \\
\noalign{\smallskip} \hline    \noalign{\smallskip}  \vspace{-0.25cm} \\ 
\rotatebox{90}{\hspace{1cm} \textbf{SF-RNN}}  & \includegraphics[width=0.99\textwidth]{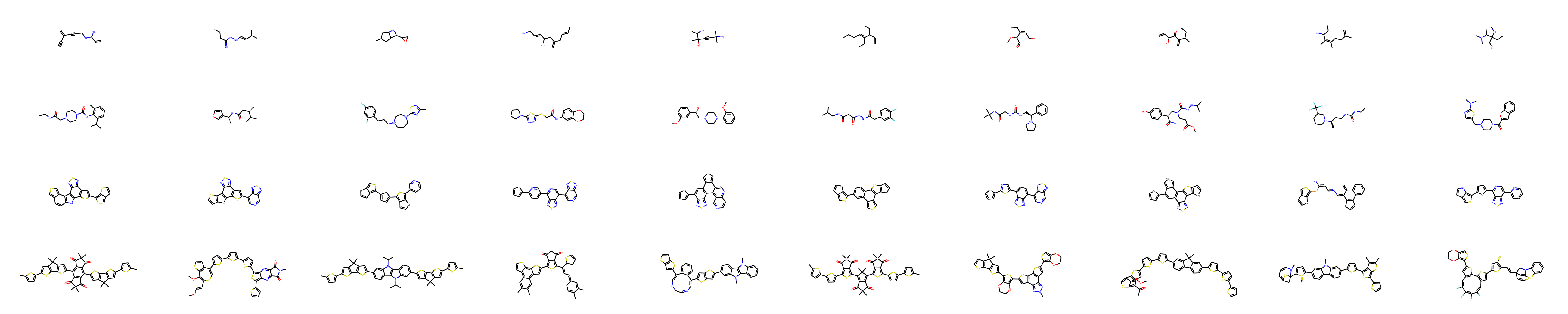} \\
\noalign{\smallskip} \hline    \noalign{\smallskip}  \vspace{-0.25cm} \\ 
\rotatebox{90}{\hspace{1cm} \textbf{SM-RNN}}  & \includegraphics[width=0.99\textwidth]{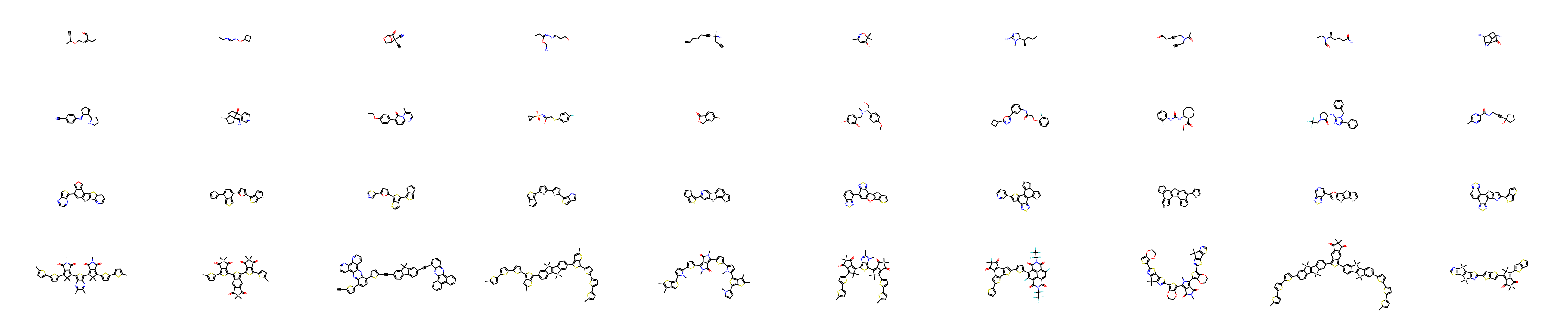} 
\end{tabular}
\caption{\textbf{Multi-distribution Task} Model generated molecules. Each sub-row is from a specific molecular mode.  }
\label{fig:mds}
\end{table*}

\begin{figure}[h]
    \centering
    \includegraphics[width=0.9\textwidth]{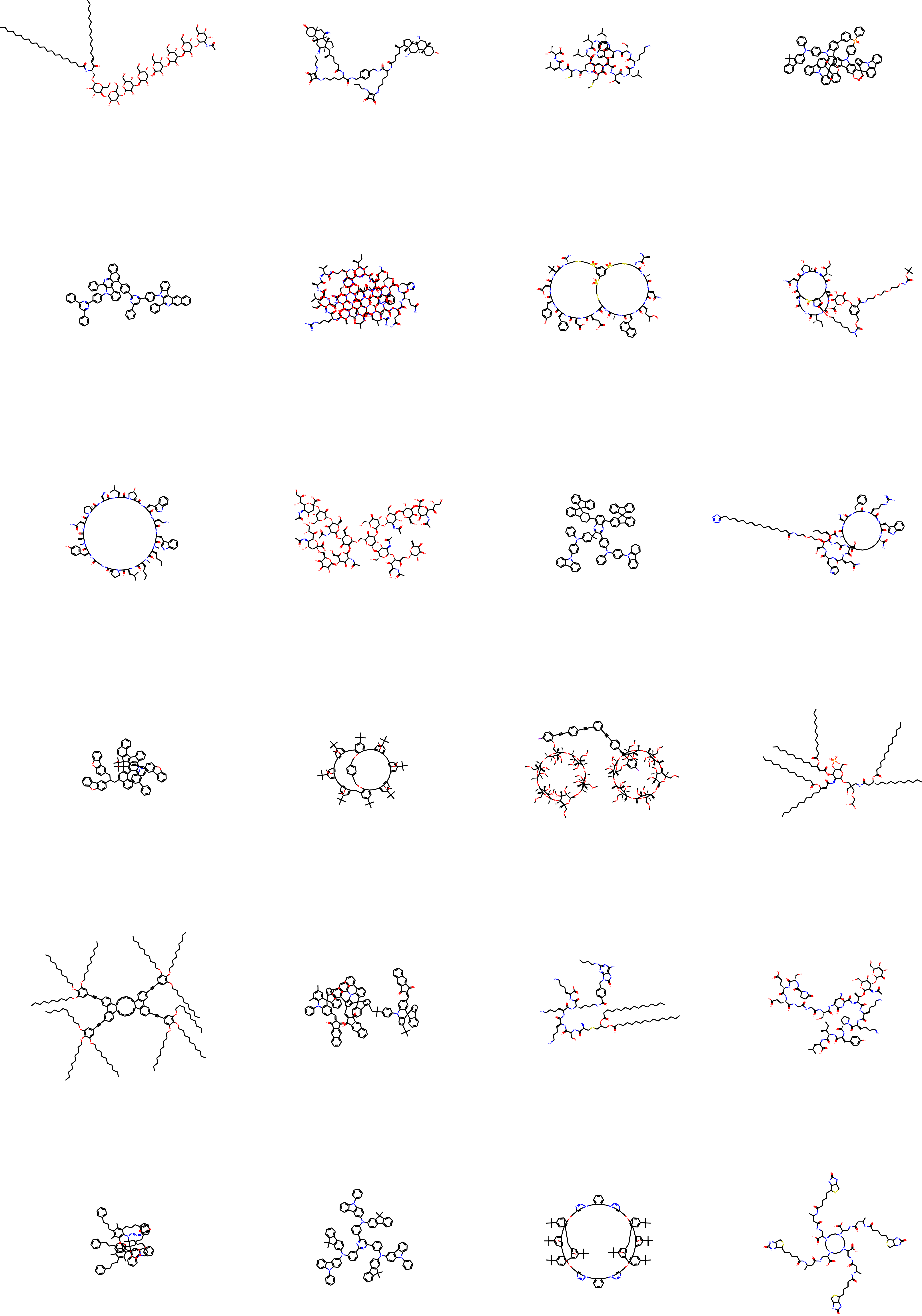}
    \caption{\textbf{Large Scale Task} Training molecules.}
    \label{fig:largetrain}
\end{figure}

\begin{figure}[h]
    \centering
    \includegraphics[width=0.9\textwidth]{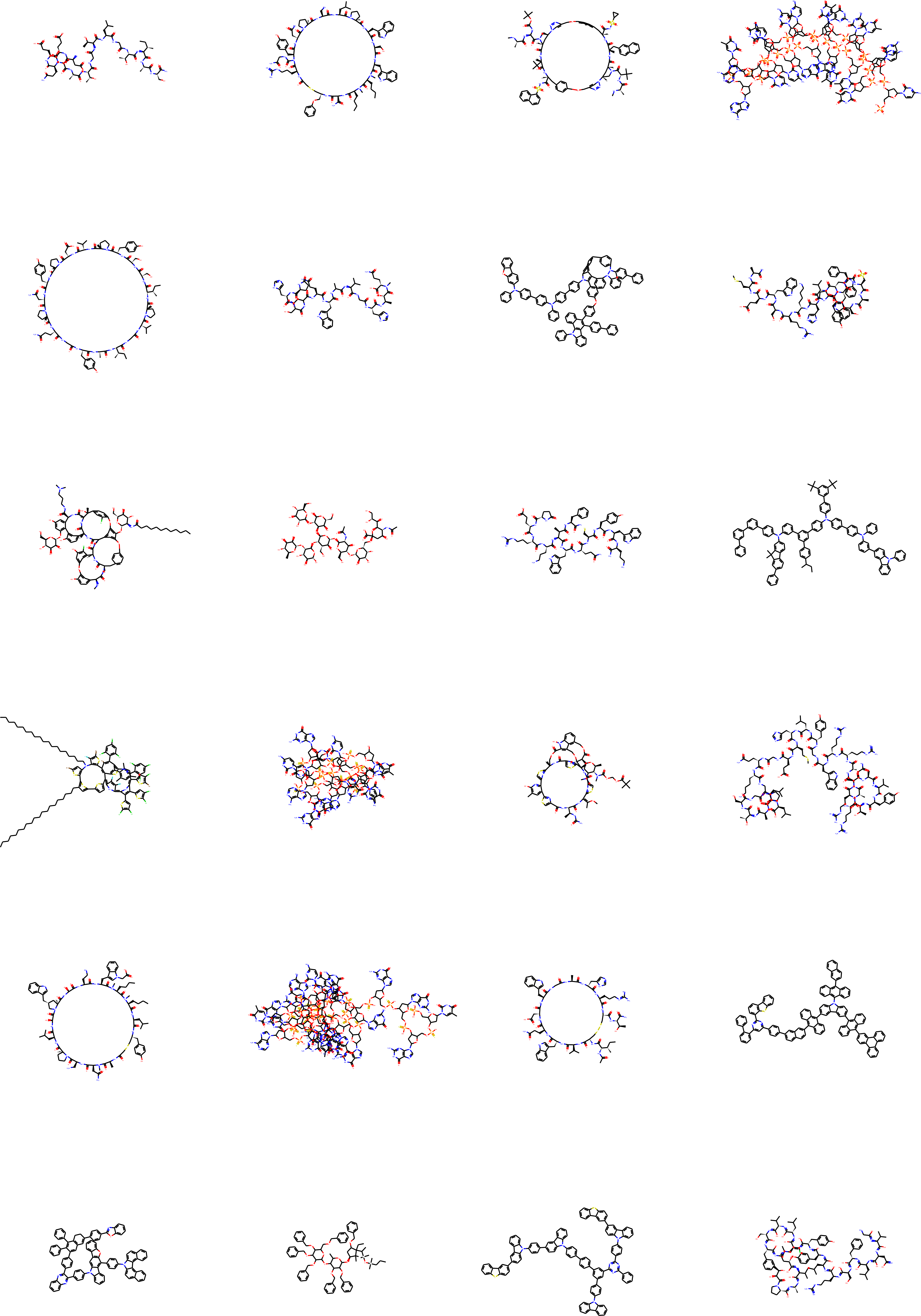}
    \caption{\textbf{Large Scale Task} Generated molecules from the SMILES RNN.}
    \label{fig:largernn}
\end{figure}  

\begin{figure}[h]
    \centering
    \includegraphics[width=0.9\textwidth]{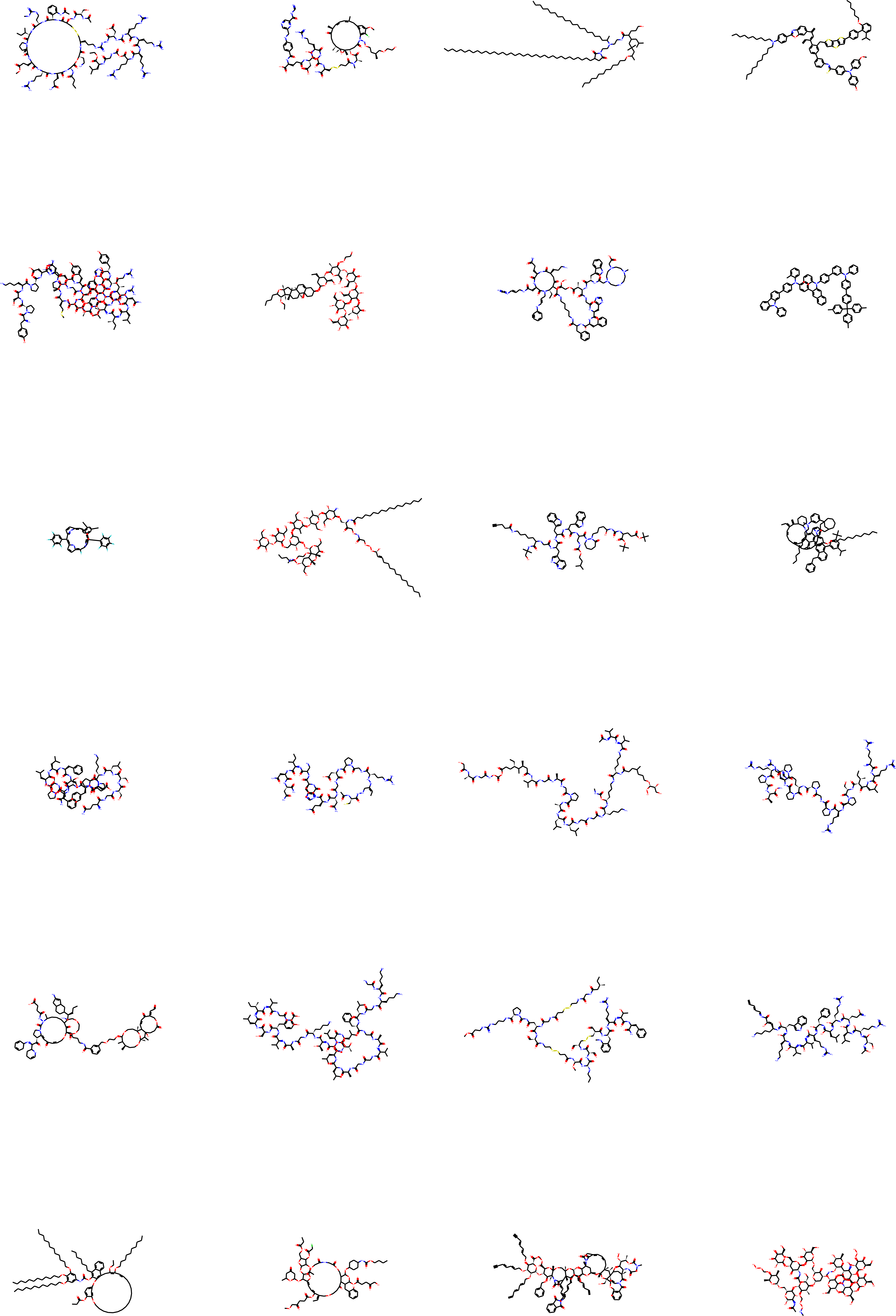}
    \caption{\textbf{Large Scale Task} Generated molecules from the SELFIES RNN.}
    \label{fig:largesfrnn}
\end{figure}  

\begin{figure}[h]
    \centering
    \includegraphics[width=0.95\textwidth]{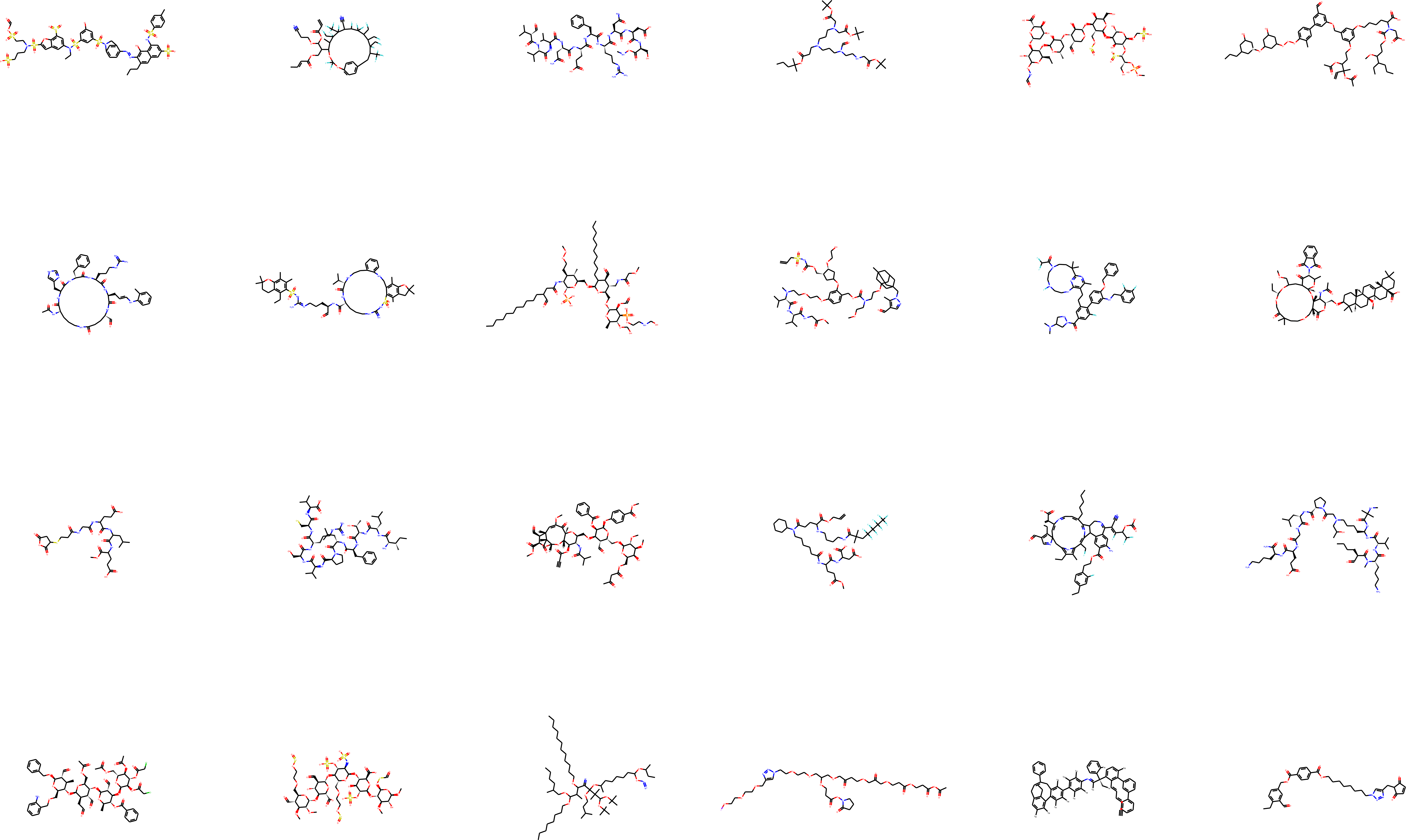}
    
    \vspace{0.25cm}

    \textbf{a } Molecules from the SELFIES RNN. 
    
     \includegraphics[width=0.95\textwidth]{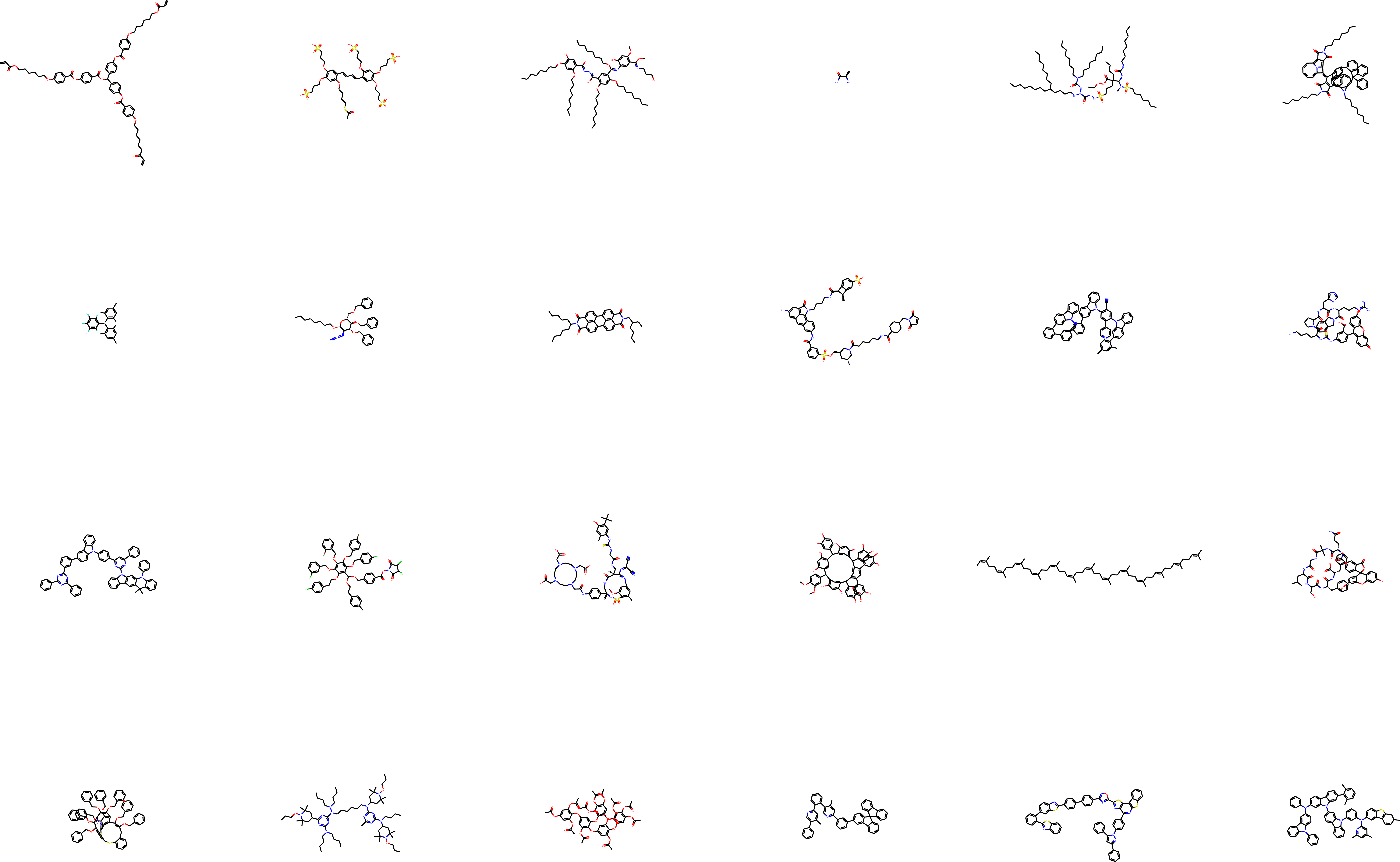}
    
    \vspace{0.25cm}

    \textbf{b } Molecules from the SMILES RNN. 
    
    \caption{\textbf{Large Scale Task} \textbf{a-b} Generated molecules with less than 100 heavy atoms from the RNN models.}
    \label{fig:lsood}
\end{figure}

\begin{figure}[h]
    \centering
    \includegraphics[width=0.95\textwidth]{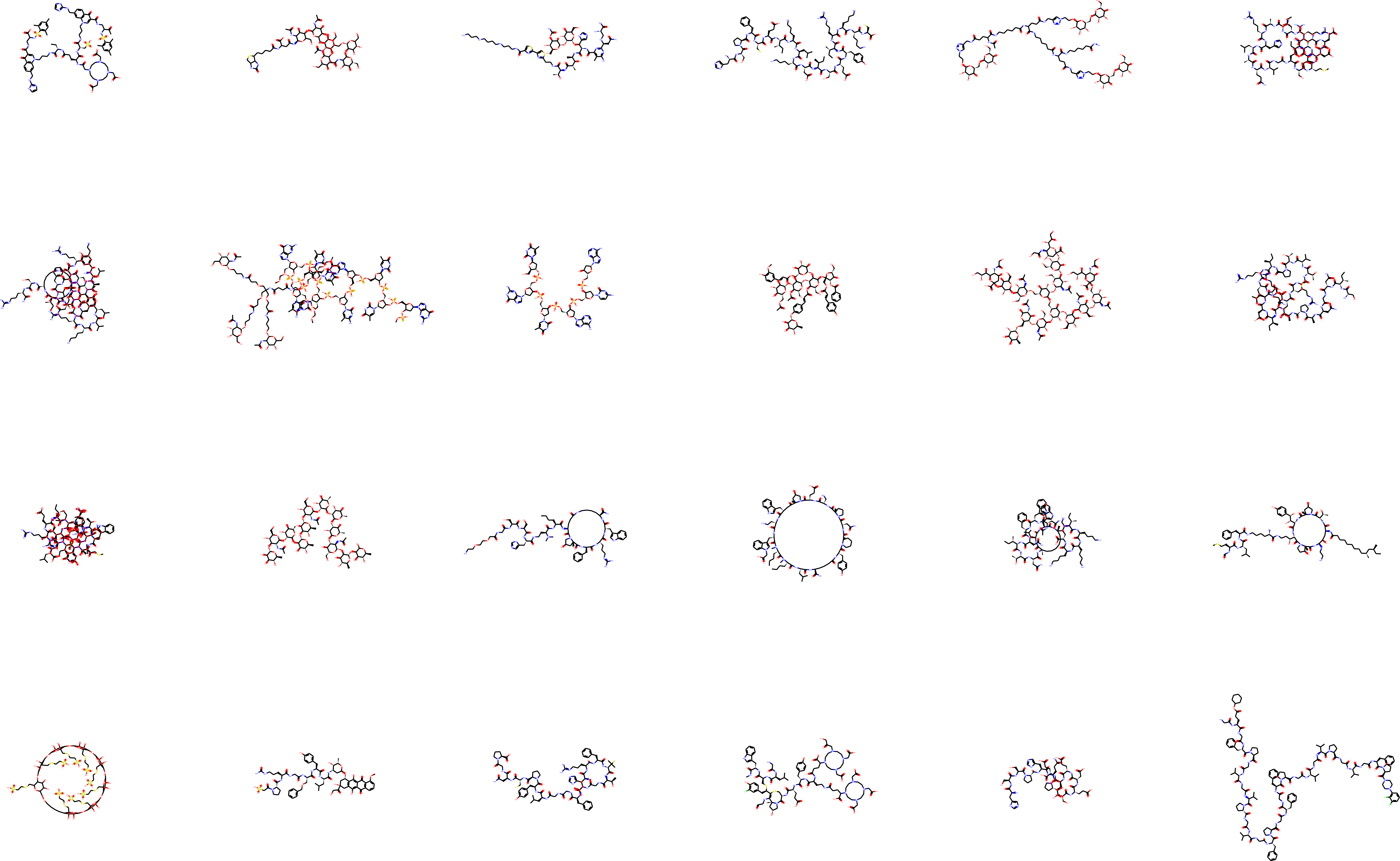}
    
    \vspace{0.25cm}

    \textbf{a } Molecules from the mode with lower LogP values.
    
    \vspace{0.25cm}
    
     \includegraphics[width=0.95\textwidth]{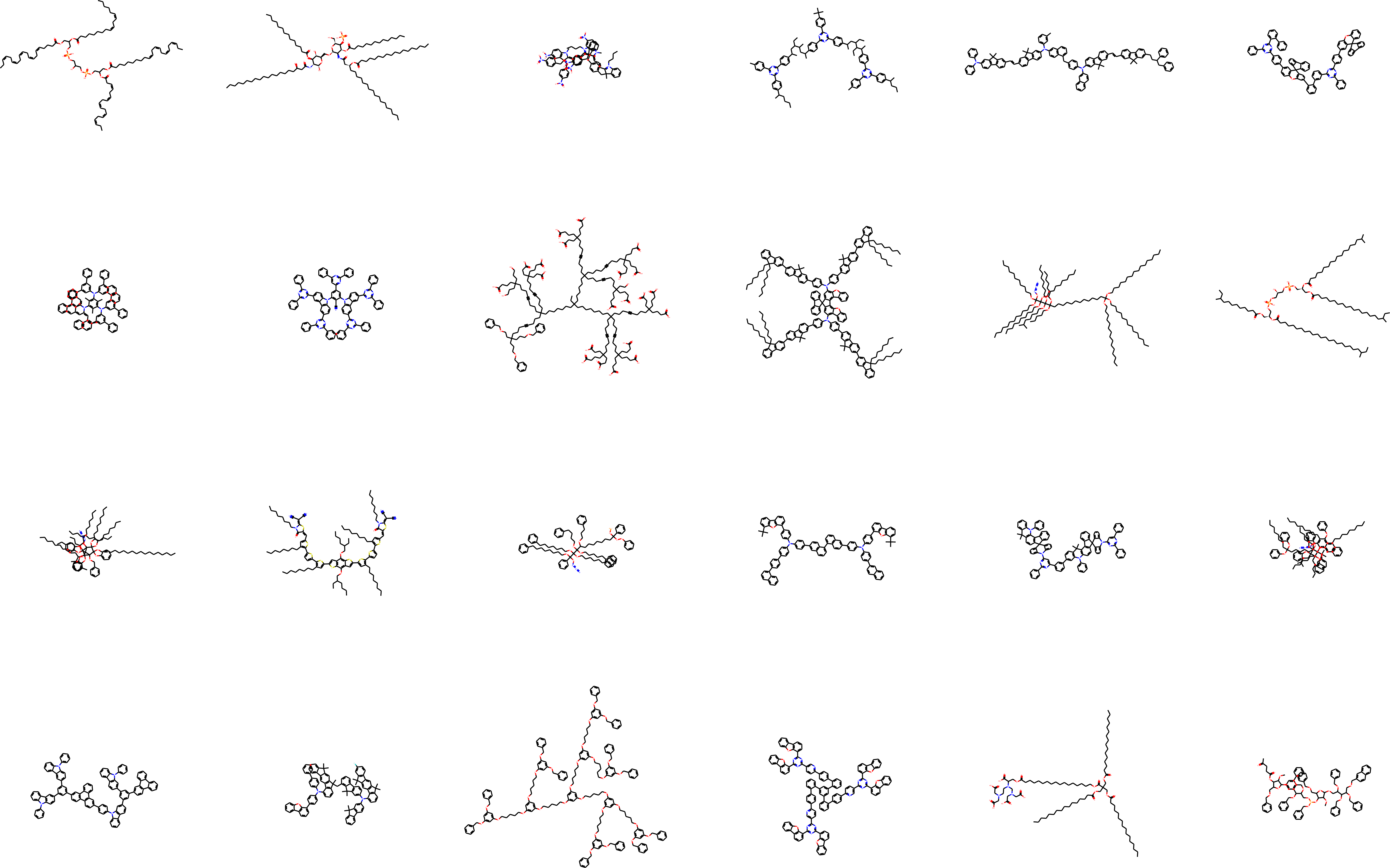}
    
    \vspace{0.25cm}

    \textbf{b } Molecules from the mode with higher LogP values.
    
    \vspace{0.25cm}
    
    \caption{\textbf{Large Scale Task} \textbf{a-b} Training molecules from each LogP mode.}
    \label{fig:largelogp}
\end{figure}

\end{widetext}

\end{document}